%% file: main.tex
\documentclass[twoside,11pt]{article}
\input{packages.tex}

\ShortHeadings{Causal discovery in the presence of missing data}{Tu, Zhang, Ackermann, Bertilson, Glymour, Kjellstr\"om and Zhang}
\firstpageno{1}

\begin{document}

\title{ Causal discovery in the presence of missing data}

\author{\name Ruibo Tu * \email ruibo@kth.se \\
       \addr KTH Royal Institute of Technology\\
       Stockholm, Sweden
       \AND
      \name Kun Zhang * \email kunz1@cmu.edu \\
       \addr Carnegie Mellon University\\
       Pittsburgh, PA, USA
       \AND
       \name Paul Ackermann \email paul.ackermann@sll.se\\
       \addr Karolinska Institute\\
       Stockholm, Sweden
       \AND
       \name Bo Christer Bertilson \email bo.bertilson@ki.se \\
       \addr Karolinska Institute\\
       Stockholm, Sweden
       \AND
       \name Clark Glymour \email cg09@andrew.cmu.edu\\
       \addr Carnegie Mellon University\\
       Pittsburgh, PA, USA
       \AND
       \name Hedvig Kjellstr\"om  \email hedvig@kth.se\\
       \addr KTH Royal Institute of Technology\\
       Stockholm, Sweden
       \AND
       \name Cheng Zhang * \email Cheng.Zhang@microsoft.com \\
       \addr Microsoft Research\\ 
       Cambridge, UK}
\maketitle

\begin{abstract}
Missing data are ubiquitous in many domains including healthcare. When these data entries are not missing completely at random, the (conditional) independence relations in the observed data may be different from those in the complete data generated by the underlying causal process. Consequently, simply applying existing causal discovery methods to the observed data may lead to  wrong conclusions. In this paper, we aim at developing a causal discovery method to recover the underlying causal structure from observed data that follow different missingness mechanisms, including missing completely at random (MCAR), missing at random (MAR), and missing not at random (MNAR). With missingness mechanisms represented by missingness graphs, we analyse conditions under which additional correction is needed to derive conditional independence/dependence relations in the complete data. 
Based on our analysis, we propose the Missing Value PC (MVPC) algorithm for both continuous and binary variables, which extends the PC algorithm to incorporate additional corrections. 
Our proposed MVPC is shown in theory to give asymptotically correct results even on data that are MAR or MNAR. 
Experimental results on synthetic data show that the proposed algorithm is able to find correct causal relations even in the general case of MNAR. Moreover, we create a neuropathic pain diagnostic simulator for evaluating causal discovery methods. Evaluated on such simulated neuropathic pain diagnosis records and the other two real world applications, MVPC outperforms the other benchmark methods\footnote{The implementation of MVPC is available at \href{https://github.com/TURuibo/MVPC}{https://github.com/TURuibo/MVPC}.}.
\end{abstract}


\input{intro.tex}
\input{related_work.tex}
\input{problem_setting.tex}
\input{method.tex}
\input{experiments.tex}
\input{discussion.tex}

\input{conclusion.tex}
\newpage
\bibliography{ref}
\end{document}

%% file: packages.tex
\usepackage{jmlr2e}

\usepackage[utf8]{inputenc} 
\usepackage[T1]{fontenc}    
\usepackage{hyperref}       
\usepackage{url}            
\usepackage{booktabs}       
\usepackage{amsfonts}       
\usepackage{nicefrac}       
\usepackage{algpseudocode}
\usepackage{algorithm}
\usepackage{microtype}      
\usepackage{amsmath}
\usepackage{graphicx}
\usepackage{pgfplots}

\usepackage{tikz}
\usetikzlibrary{patterns}
\usetikzlibrary{arrows, automata}
\usepackage{filecontents}
\usepackage{amsthm}
\usepackage{paralist, tabularx}
\usepackage{microtype}
\usepackage{textcase}

\usepackage{stmaryrd}

\usepackage{subcaption} 
\usepackage{enumitem}
\usepackage{wrapfig}

\usepackage{longtable,calc}
\usepackage{mathptmx}       
\usepackage[textwidth=2cm,colorinlistoftodos]{todonotes}
\usepackage{xspace}
\usepackage{latexsym}
\usepackage{times}   
\usepackage{multirow}
\usepackage{color,comment,rotating,url,xspace} 
\usetikzlibrary{arrows,shadows,shapes,backgrounds,decorations,snakes,fit}
\usepackage[verbose=true,letterpaper]{geometry}
\usepackage{mathtools}

\makeatletter
\newcommand*{\centernot}{%
  \mathpalette\@centernot
}
\def\@centernot#1#2{%
  \mathrel{%
    \rlap{%
      \settowidth\dimen@{$\m@th#1{#2}$}%
      \kern.5\dimen@
      \settowidth\dimen@{$\m@th#1=$}%
      \kern-.5\dimen@
      $\m@th#1\not$%
    }%
    {#2}%
  }%
}
\makeatother
\newcommand{\independent}{\,\perp\mkern-9.5mu\perp\,}
\newcommand{\spacevert}{\mid}
\newcommand{\notindependent}{\centernot{\independent}}
\newcommand{\dsep}{\independent _{\mkern-9.5mu d}}
\newcommand{\notdsep}{\notindependent _{\mkern-9.5mu d}}

\newtheorem{theorem}{Theorem}

\newtheorem{proposition}[theorem]{Proposition}

\theoremstyle{assumption}
\newtheorem{assumption}{Assumption}

\pgfdeclarepatternformonly{south west lines}{\pgfqpoint{-0pt}{-0pt}}{\pgfqpoint{3pt}{3pt}}{\pgfqpoint{3pt}{3pt}}{
        \pgfsetlinewidth{0.4pt}
        \pgfpathmoveto{\pgfqpoint{0pt}{0pt}}
        \pgfpathlineto{\pgfqpoint{3pt}{3pt}}
        \pgfpathmoveto{\pgfqpoint{2.8pt}{-.2pt}}
        \pgfpathlineto{\pgfqpoint{3.2pt}{.2pt}}
        \pgfpathmoveto{\pgfqpoint{-.2pt}{2.8pt}}
        \pgfpathlineto{\pgfqpoint{.2pt}{3.2pt}}
        \pgfusepath{stroke}}
\pgfplotsset{compat=1.11,
        /pgfplots/ybar legend/.style={
        /pgfplots/legend image code/.code={%
        \draw[##1,/tikz/.cd,bar width=12pt,yshift=-0.2em,bar shift=0pt]
                plot coordinates {(0cm,0.8em)};},
},
}        
\makeatletter

\newcommand{\hk}[1]{{\color{black} #1}} 



%% file: intro.tex
\section{Introduction} \label{sec:intro}
Determining causal relations plays a pivotal role in many disciplines of science, especially in healthcare. In particular, understanding causality in healthcare can facilitate effective treatments to improve quality of life. Traditional approaches  \citep{domeij2016ageing} to identify causal relations are usually based on randomized controlled trails, which are expensive or even impossible in certain domains. In contrast, owing to the availability of purely observational data and recent technological developments in computational and statistical analysis, causal discovery from observational data is potentially widely applicable ~\citep{Spirtes01,Pearl00}. In recent years, causal discovery from observational data has become popular in medical research \citep{sokolova2017handling, klasson2017causality}.

Most existing algorithms for causal discovery are designed for complete data \citep{Pearl00,  peters2017elements}, such as the widely used PC algorithm~\citep{Spirtes01}. Unfortunately, missing data entries are  common in many domains. For example, in healthcare, missing entries may come from imperfect data collection, compensatory medical instruments, and fitness of the patients etc. \citep{robins1986new}.

All missing data mechanisms fall into one of the following three categories \citep{rubin1976inference}: Missing Completely At Random (MCAR), Missing At Random (MAR), and Missing Not At Random (MNAR).  Data are MCAR if the cause of missingness is purely random, e.g., some entries are deleted due to a random computer error. Data are MAR when the direct cause of missingness is fully observed. For example, a dataset consists of two variables: gender and income, where gender is always observed and income has missing entries. MAR missingness would occur when men are more reluctant than women to disclose their income (i.e., gender causes missingness).  Data that are neither MAR nor MCAR fall under the MNAR category. In the example above, MNAR would occur when gender also has missing entries. These missingness mechanisms can be represented by causal graphs as introduced in Section \ref{sec:related}. While it might be tempting to remove samples corrupted by missingness and perform analysis solely with complete cases, it will reduce sample size and, more importantly, bias the outcome especially when data are MAR or MNAR \citep{rubin2004multiple, mohan2013graphical}. 

This paper is concerned with how to find the underlying causal structure from observed data even in the situation of MAR or MNAR. For simplicity, we assume causal sufficiency in the paper, as assumed by many causal discovery methods including PC \citep{Spirtes01}. Recoverability of the data distribution under missing data has been discussed in a number of contributions; see, e.g., \citep{mohan2013graphical,mohan2014graphical}. A straightforward solution is to recover all relevant distributions that are needed for Conditional Independence (CI) tests involved in the CI-based causal search procedure, such as PC. But compared to the CI test on independent and identically distributed observations, the CI test implied by corrected distributions is generally harder to assess because it involves simulating new data or importance reweighting with density ratios. 
Therefore, instead of correcting all CI tests of the PC algorithm, we aim to find under which conditions CI tests in the observed data produce erroneous edges, and subsequently correct only such edges by further applying CI tests on corrected distributions. 
Our main contributions are: 
\begin{compactitem}

\item \emph{ We provide a theoretical analysis of the error that different missingness mechanisms introduce in the results given by traditional causal discovery methods, especially the PC algorithm (Section \ref{sec:problem_setting}).} 
We show that naive deletion-based method may lead to incorrect results due to the bias caused by missing data. 
One immediate way to extend constraint-based methods to handle the missing data issue is correcting all the involved CI tests. This approach is neither data-efficient nor computation-efficient. Therefore, we identify possible errors that different missingness mechanisms lead to in the results given by deletion-based PC. We show that one usually needs to correct only a small number of CI tests in order to recover the true causal structure.

\item \emph{We propose a novel, correction-based extension of the PC algorithm, Missing Value PC (MVPC), that handles all three types of missingness mechanisms: MCAR, MAR, and MNAR (Section \ref{sec:method}). }
Based on the result from Section \ref{sec:problem_setting}, we identify  where corrections are required and propose efficient correction methods for all three types of the missingness mechanisms.

\item  \emph{We show the superior performance of MVPC in different settings, including three real-life healthcare scenarios (Section \ref{sec:exp}).}
We first evaluate the proposed MVPC on synthetic datasets under different settings. MVPC shows clear improvement over multiple baselines. 
We then develop a neuropathic pain diagnosis simulator for evaluating various types of causal discovery algorithms, including those to deal with practical issues in causal discovery, such as unknown confounders, selection bias, and missing data.
We further apply MVPC to our neuropathic pain diagnosis simulation datasets and two real-world datasets in the US Cognition study and Achilles Tendon Rupture study. The results are consistent with medical domain knowledge and show the efficacy of our method.
\end{compactitem}

%% file: related_work.tex
\section{Related work}
\label{sec:related}

We discuss closely related works, including traditional causal discovery algorithms and approaches that deal with missing data from a causal perspective.

\paragraph{Causal discovery.}
Causal discovery from observational data has been of great interest in various domains in the past decades \citep{Pearl00,Spirtes01}. In general, causal discovery consists of constraint-based methods, score-based methods, and methods based on functional causal models. Typical constraint-based methods include the PC algorithm and Fast Causal Inference (FCI) \citep{Spirtes01}. They assume that all CI relations in the data are entailed by the causal Markov condition on the underlying causal graph, known as the faithfulness assumption, and use CI constraints in the data to recover causal structure. 
The PC algorithm  assumes no confounders (i.e., hidden direct common causes of two variables) and outputs a Completed Partially Directed Acyclic Graph (CPDAG), which represents the Markov equivalence class that contains all DAGs that have the same CI relations, and is easy to interpret and often used in biomedical applications \citep{neto2008inferring, le2016fast}. FCI allows confounders and selection bias, and outputs a Partial Ancestral Graph (PAG). For simplicity, we use the PC algorithm in this paper, but it is straightforward to transfer our framework to other constraint-based methods. Score-based methods (e.g., Greedy Equivalence Search \citep{chickering2002optimal}) find the best Markov equivalence class under certain score-based criterion, such as the Bayesian Information Criterion (BIC).
Causal discovery based on functional causal models benefits from the additional assumptions on the data distribution and/or the functional classes to further determine the causal direction between variables. Typical functional causal models include the linear non-Gaussian acyclic model (LiNGAM) \citep{shimizu2006linear}, the post-nonlinear (PNL) causal model \citep{zhang2009identifiability,Zhang06_iconip}, and the nonlinear additive noise model (ANM) \citep{peters2017elements}.

\paragraph{Dealing with data with missing values from a causal perspective.}
Recent years have witnessed a growing interest in analysing the problem of missing data from a causal perspective.  In particular, the notions of recoverability and testability have been studied by modeling the missingness process using causal graphs (called missingness graphs)  \citep{mohan2013graphical}. Given a missingness graph, a query (such as conditional or joint distribution and causal effects) is deemed recoverable if it can be consistently estimated  \citep{mohan2014graphical}. With the application of missingness graphs, \citet{shpitser2015missing} represent missing data as a causal inference problem and identify joint distributions in MNAR settings. Relaxing the assumption of \citep{shpitser2015missing}, \citet{rozi19mid} propose a method to identifying a wider class of queries. Moreover, \citet{saadati2019adjustment} use missingness graphs, provide the conditions under which the causal effect in the presence of MNAR can be recovered, and propose an algorithm to find valid variable sets for the recovery. Testability, on the other hand, deals with finding testable implications, i.e., claims refutable by the (missing) data distribution \citep{mohan2014testability}. As for causal discovery in the presence of missing values, it aims to find the structure of variables of interest rather than the missingness. Under appropriate assumptions, relations of concerned variables can be testable.


In causal discovery, there are few works for the MNAR case. \citet{strobl2017fast} regard the missingness procedure as a particular type of selection bias to handle the MNAR missingness and perform FCI by test-wise deletion. The test-wise deletion removes the incomplete records of variables involved in each CI test. It shows that FCI combined with test-wise deletion is still sound when one aims to estimate the PAG for the variables including the effect of missingness. Data missingness is usually different from selection bias, because in the selection bias case we only have the distribution of the selected samples but no clue about the population. However, in the missing data case, we may be able to check the (conditional) independence relation between two variables given others by making use of the available data for the involved variables. In the case where the missingness mechanisms are known, this problem is closely related to  recoverability of models with missing data. \citet{gain2018structure} utilize Inverse Probability Weight (IPW) for each CI test, assuming the missing data model is known, which may not be realistic in many real-life applications. When the missing data model is unknown, they choose the sparest resulting graph considering all possible missingness structures, which is usually computationally expensive. 

%% file: problem_setting.tex
\section{Deletion-based PC: A first proposal and its behavior}\label{sec:test-w-del}\label{sec:problem_setting}

We assume that there is no confounder or selection bias relative to the set of observed variables. When the available dataset has missing values, one may apply the PC algorithm for causal discovery by performing CI tests on those records which do not have missing values for the variables involved in the tests. We term this first proposal \emph{deletion-based PC}. In this section, we discuss the influence of missing data on the result of deletion-based PC.

Primarily, we investigate the situations where errors occur to the output of deletion-based PC due to the missingness. Firstly, we utilize missingness graphs and summarize the assumptions that we need for properly dealing with missingness. We then present the aforementioned deletion-based PC algorithm. 
Our analysis focuses on properties of the results given by this naive extension, and provides the conditions under which the deletion-based PC produces erroneous edges.

\paragraph{Missingness graph.} 
We utilize the notation of the  \emph{missingness graph} \citep{mohan2013graphical}. 
A missingness graph is a causal DAG $G(\mathbb{V} ,\mathbf{E})$ where $\mathbb{V} = \mathbf{V} \cup \mathbf{U} \cup \mathbf{V^*} \cup \mathbf{R}$. $\mathbf{U}$ is the set of unobservable nodes; in this paper, we assume causal sufficiency, so $\mathbf{U}$ is an empty set. $\mathbf{V}$ is the set of \textit{substantive} nodes (observable nodes) containing $\mathbf{V}_o$ and $\mathbf{V}_m$. $\mathbf{V}_o \subseteq \mathbf{V}$ is the set of fully observed variables, denoted by white nodes in our graphical representation. $\mathbf{V}_m \subseteq \mathbf{V}$ is the set of partially observed variables that are missing in at least one record, which is shadowed in gray. $\mathbf{R}$ is the set of \textit{missingness indicators} that represent the status of missingness and are responsible for the values of proxy variables $\mathbf{V^*}$. For example, the proxy variable $Y^* \in \mathbf{V^*}$ is introduced as an auxiliary variable for the convenience of derivation. $R_y=1$ means that the corresponding record value of $Y$ is missing and $Y^*$ corresponds to a missing entry; $R_y=0$ indicates that the corresponding record value of $Y$ is observed and $Y^*$ takes the value of $Y$.

In this work we adopt the CI-based definitions of missingness categories as stated  in \citep{mohan2013graphical}. We denote an independence relation in a dataset by "$\independent$" and d-separation in a missingness graph by "$\dsep$". As shown in Figure \ref{fig:con_est}, data are MCAR if $\{\mathbf{V}_m,\mathbf{V}_o\} \dsep \mathbf{R}$ holds in the missingness graph, MAR if $\mathbf{V}_m \dsep \mathbf{R} \mid \mathbf{V}_o$ holds, and MNAR otherwise.  
\begin{figure*}[!ht]%
\centering
\begin{subfigure}{.2\textwidth}
  \centering
  \input{tikz/MCAR_example.tex}
 \caption{A MCAR graph}
  \label{fig:mcar}
\end{subfigure}%
\hspace{4mm}
\begin{subfigure}{.2\textwidth}
  \centering
    \input{tikz/MAR_example.tex}
  \caption{A MAR graph}
  \label{fig:mar}
\end{subfigure}
\hspace{4mm}
\begin{subfigure}{.2\textwidth}
	\centering
	\input{tikz/determ_r.tex}  
    \caption{A MNAR graph} 
    \label{fig:mnar}
\end{subfigure}
\hspace{4mm}
\begin{subfigure}{.2\textwidth}
  \centering
    \input{tikz/self_missing1.tex}
  \caption{Self-masking missingness}
  \label{fig:self-m-sing}
\end{subfigure}
\caption{Exemplar missingness graphs in MCAR, MAR, MNAR, and self-masking missingness. $X$, $Y$, $Z$, and $W$ are random variables. In missingness graphs, gray nodes are partially observed variables, and white nodes are fully observed variables. $R_x$, $R_y$, and $ R_w$ are the missingness indicators of $X$, $Y$, and $W$.}
\label{fig:con_est}
\end{figure*}
\paragraph{Assumptions for dealing with missingness.} 
Apart from the assumptions for the asymptotic correctness of the PC algorithm (including the causal Markov condition, faithfulness, and no confounding or selection bias), we introduce some additional assumptions that we make use of to deal with missingness.
\begin{assumption}[Missingness indicators are not causes]\label{ass:re0}
No missingness indicator can be a cause of any substantive (observed) variable. 
\end{assumption}
This assumption is employed in most related work using missingness graphs \citep{mohan2013graphical, mohan2014graphical}. Consequently, under this assumption, if variables of interest $X$ and $Y$ are not d-separated by a variable set $\mathbf{Z} \subseteq \mathbf{V} \setminus \{X,Y\}$, they are not d-separated by $\mathbf{Z}$ together with their missingness indicators. Under the faithfulness assumption, this means that if they are conditionally independent given $\mathbf{Z}$ together with the their missingness indicators, they are conditionally independent given only $\mathbf{Z}$. 
Now the problem is that generally speaking, we cannot directly verify whether they are conditionally independent given $\mathbf{Z}$ and their missingness variables because we do not have the records for the considered variables when their missingness indicators take value one. We then need the following assumptions to deal with this issue.
\begin{assumption}[Faithful observability] 
Any conditional independence relation in the observed data also holds in the unobserved data; formally, $\,X \independent Y \spacevert \{\mathbf{Z}, \mathbf{R_K}=\mathbf{0}\} \Longleftrightarrow \,X \,\independent Y \spacevert \{\mathbf{Z}, \mathbf{R_K}=\mathbf{1}\}$. Here $\mathbf{R_K}$ is the missingness indicator set $\{R_x, R_y, \mathbf{R_z}\}$. $\mathbf{R_K}=\mathbf{0}$ means that all the missingness indicators in $\mathbf{R_K}$ take the value zero; $\mathbf{R_K}=\mathbf{1}$ means that at least one missingness indicator in $\mathbf{R_K}$ takes the value one. \end{assumption}

This implies $X \independent Y \spacevert \{\mathbf{Z}, \mathbf{R_K}=\mathbf{0}\} \Longleftrightarrow X \independent Y \spacevert \{\mathbf{Z}, \mathbf{R_K}\}$, which means that conditional independence relations in the observed data also hold in the complete data, i.e., there is no accidental conditional independence relation caused by missingness.  

\begin{assumption}[No deterministic relation between missingness indicators]\label{ass:re1}
No missingness indicator can be a deterministic function of any other missingness indicators. 
\end{assumption}

\begin{assumption}[No self-masking missingness]  \label{ass:re2}
Self-masking missingness refers to missingness in a variable that is caused by itself. In the missingness graph this is depicted by an edge from $X$ to $R_x$, for $X \in \mathbf{V_m}$, as shown in Figure \ref{fig:self-m-sing}. We assume that there is no such edges in the missingness graph. 
\end{assumption}

Assumption \ref{ass:re1} and \ref{ass:re2} guarantee the recoverability of a joint distribution of substantive variables, as shown in \citep{mohan2013graphical}. As discussed in Section \ref{sec:discussion}, in linear Gaussian cases the "self-masking" only affects causal discovery results when $R_x$ has direct causes other than $X$.

In the end, we assume linear Gaussian causal models in the continuous case.
Thus, one can check CI relations with the partial correlation test, a simple CI test method. 
Note that our proposed algorithm also works well for other scenarios. 
In the binary case, we can simply use the $G^2$ test.
In the non-linear case, we can use a suitable non-linear or non-parametric one \citep{Zhang11_KCI}. 

\paragraph{Effect of missing data on the deletion-based PC.}
In the presence of missing data, the list-wise deletion PC algorithm deletes all records that have any missing value and then applies the PC algorithm to the remaining data. In contrast, the test-wise deletion PC algorithm only deletes records with missing values for variables involved in the current CI test when performing the PC algorithm (which can be seen as the PC algorithm realization of \citep{strobl2017fast}). Test-wise deletion is more data-efficient than list-wise deletion. In this paper, we focus on the Test-wise Deletion PC algorithm (TD-PC). 

TD-PC gives asymptotically correct results when data are MCAR since $\{\mathbf{V}_m,\mathbf{V}_o\} \dsep \mathbf{R}$ is satisfied. Consider Figure \ref{fig:mcar} as an example. $R_y \dsep \{X,Y,Z\}$ holds; thus, we have $X \dsep Y \spacevert Z \,\Longleftrightarrow \;X \dsep Y \spacevert \{Z, R_y\}$. With the faithfulness assumption on missingness graphs, $X \independent Y \spacevert Z \Longleftrightarrow X \independent Y \spacevert \{Z, R_y\}$. Furthermore, with the faithful observability assumption, we conclude $X \independent Y \spacevert Z \Longleftrightarrow X \independent Y^* \spacevert \{Z, R_y=0\}$. 
When applying the CI test to the test-wise deleted data of concerned variables $X$, $Y$, and $Z$, we test whether $X \independent Y^* \mid \{Z,R_y=0\}$ holds.
Therefore, CI results imply d-separation/d-connection relations of concerned variables in missingness graphs when data are MCAR, which guarantees the asymptotic correctness of TD-PC.

In cases of MAR and MNAR, TD-PC may produce erroneous edges because $\{\mathbf{V}_m,\mathbf{V}_o\} \dsep \mathbf{R}$ does not hold.
Therefore, in what follows in this section, we mainly address the problems of TD-PC in cases of MAR and MNAR.

\paragraph{Erroneous edges produced by TD-PC.}\label{sec:id}
Since TD-PC may produce erroneous edges when data are MAR and MNAR, in the following propositions, we first show that the causal skeleton (undirected graph) given by TD-PC has no missing edges, but may contain extraneous edges. We then determine the conditions under which extraneous edges occur in the output of TD-PC. 
\begin{proposition}\label{prop:extra_edges}
Under Assumptions \ref{ass:re0}$\sim$\ref{ass:re2}, the CI relation in test-wise deleted data, $X \independent Y \spacevert \{\mathbf{Z},R_x=0,R_y=0,\mathbf{R_z}=\mathbf{0}\}$, implies the CI relation in complete data,  $X \independent Y \spacevert \mathbf{Z}$, where $X$ and $Y$ are random variables and $\mathbf{Z}\subseteq \mathbf{V} \setminus \{X,Y\}$.
\end{proposition}
\begin{proof} 
$X\independent Y|\{\mathbf{Z},\mathbf{R_z}=\mathbf{0}, R_x=0, R_y=0\}
\Rightarrow
X\independent Y|\mathbf{Z}$: 
We have $X\independent Y|\{\mathbf{Z},\mathbf{R_z}=\mathbf{0}, R_x=0, R_y=0\}$, where some of the involved missingness indicators may only take value 0 (i.e., the corresponding variables do not have missing values). With the faithful observability assumption, this condition implies $X\independent Y|\{\mathbf{Z},\mathbf{R_z}, R_x, R_y\}$. Because of the faithfulness assumption on missingness graphs, we know that $X$ and $Y$ are d-separated by $\{\mathbf{Z},\mathbf{R_z}, R_x, R_y\}$; furthermore, according to Assumptions \ref{ass:re0}, \ref{ass:re1}, and \ref{ass:re2}, the missingness indicators can only be leaf nodes in the missingness graph. Therefore, conditioning on these nodes will not destroy the above d-separation relation. That is, in the missingness graph, $X$ and $Y$ are d-separated by $\mathbf{Z}$. Hence, we have  $X\independent Y\mid\mathbf{Z}$.
\end{proof}

Proposition \ref{prop:extra_edges} shows 
that CI relations in test-wise deleted data implies the true corresponding d-separation relations in a missingness graph. However, dependence relations in test-wise deleted data may imply the wrong corresponding relations in the missingness graph because $X \notindependent Y \mid \{\mathbf{Z},R_x=0,R_y=0,\mathbf{R_z}=\mathbf{0}\} \nRightarrow X \notindependent Y \mid \mathbf{Z}$. In other words, TD-PC may wrongly treat some d-separation relations of concerned variables as  d-connected (not d-separated) in a missingness graph. Thus, TD-PC produces extraneous edges in the causal skeleton result rather than missing edges. For example, in Figure \ref{fig:mar}, we have $X \notindependent Y^* \mid \{Z, R_y=0\}$ in the test-wise deleted data, but the true d-separation relation is $X \dsep Y \mid Z$ instead of $X \notdsep Y \mid Z$. Thus, TD-PC produces an extraneous edge between $X$ and $Y$.
Fortunately, such extraneous edges  appear only under special circumstances, as shown in the following proposition. 

\begin{proposition}\label{prop:iden}
Suppose that $X$ and $Y$ are not adjacent in the true causal graph and that for any variable set $\mathbf{Z}\subseteq \mathbf{V} \setminus \{X,Y\}$ such that $X \independent Y \spacevert \mathbf{Z}$, it is always the case that $X \notindependent Y \spacevert \{\mathbf{Z},R_x = 0,R_y = 0,\mathbf{R_z } = \mathbf{0}\}$. Then under Assumptions \ref{ass:re0}$\sim$\ref{ass:re2}, for at least one variable in $\{X\} \cup \{Y\}\cup \mathbf{Z}\,$, its missingness indicator is either the direct common effect or a descendant of the direct common effect of $X$ and $Y$.
\end{proposition}
\begin{proof} 
The condition of Proposition \ref{prop:iden} implies that for nodes $X$, $Y$ and any node set $\mathbf{Z} \subseteq \mathbf{V}\setminus\{X,Y\}$ in a missingness graph, 
conditioning on $\mathbf{Z}$ and missingness indicators $R_x$, $R_y$, and $ \mathbf{R_z}$, there always exists an undirected path $U$ between $X$ and $Y$ that is not blocked. Furthermore, to satisfy such constraint of $U$, at least a missingness indicator $R_i \in \{R_x,R_y,\mathbf{R_z}\}$ satisfies either one of the following two conditions: (1) $R_i$ is the only vertex on $U$; (2) A cause of $R_i$ is the only vertex on $U$ as a collider. 
In Condition (1), if $R_i$ is on $U$, it is a collider because under Assumptions \ref{ass:re0}$\sim$\ref{ass:re2}, missingness indicators are the leaf nodes in missingness graphs. Then, suppose that $R_i$ is not the only vertex on $U$, and that another node $V_j \in \mathbf{V}\setminus\{X,Y,\mathbf{Z}\}$ is also on $U$. Conditioning on $V_j$ and $R_i$, $U$ is blocked, which is not satisfied by the constraint on $U$. Thus, $R_i$ should be the only vertex on $U$. The same reason also applies to Condition (2). In summary, we conclude that under the condition of Proposition \ref{prop:iden}, there is at least one missingness indicator $R_i \in \{R_x,R_y,\mathbf{R_z}\}$ such that $R_i$ is the direct common effect or a descendant of the direct common effect of $X$ and $Y$.
\end{proof}

Proposition \ref{prop:iden} indicates that extraneous edges can be identified from the output of TD-PC. For example, in Figure \ref{fig:mar} and Figure \ref{fig:mnar}, $W$ is the direct common effect of $X$ and $Y$ and the missingness indicator $R_y$ is a descendant of $W$. Thus, the extraneous edge occurs between $X$ and $Y$ in the causal skeleton produced by TD-PC.

%% file: tikz/MCAR_example.tex
\begin{tikzpicture}[
			scale=0.4,
            > = stealth, 
            shorten > = 1pt, 
            auto,
            node distance = 3cm, 
            semithick 
        ]

        \tikzstyle{every state}=[
            draw = black,
            thick,
            minimum size = 8mm
        ]

        \node[state] (X) at  (-2,2.5){$X$};
        \node[state] (Z) at  (-0,0) {$Z$};
        \node[state,fill=gray] (Y) at  (2,2.5) {$Y$};
        \node[state] (ry) at  (4,0) {\small$R_y$};

        \path[->] (X) edge node {} (Z);
        \path[->] (Z) edge node {} (Y);
        
    \end{tikzpicture}

%% file: tikz/MAR_example.tex
\begin{tikzpicture}[
			scale=0.4,
            > = stealth, 
            shorten > = 1pt, 
            auto,
            node distance = 3cm, 
            semithick 
        ]

        \tikzstyle{every state}=[
            draw = black,
            thick,
            minimum size = 8mm
        ]

        \node[state] (X) at  (-3,2.5){$X$};
        \node[state,fill=gray] (Y) at  (3,2.5) {$Y$};
        \node[state] (Z) at  (-0,2.5){$Z$};
		\node[state] (W) at  (-0,0) {$W$};
        \node[state] (ry) at  (3,0) {\small$R_y$};

        \path[->] (X) edge node {} (W);
        \path[->] (Y) edge node {} (W);
        \path[->] (X) edge node {} (Z);
        \path[->] (Z) edge node {} (Y);
        \path[->] (W) edge node {} (ry);
        
    \end{tikzpicture}

%% file: tikz/determ_r.tex
\begin{tikzpicture}[
			scale=0.4,
            > = stealth, 
            shorten > = 1pt, 
            auto,
            node distance = 3cm, 
            semithick 
        ]

        \tikzstyle{every state}=[
            draw = black,
            thick,
            minimum size = 8mm
        ]

        \node[state] (X) at  (-3,2.5){$X$};
        \node[state,fill=gray] (W) at  (-0,0) {$W$};
        \node[state,fill=gray] (Y) at  (3,2.5) {$Y$};
        \node[state] (Z) at  (0,2.5){$Z$};

        \node[state] (ry) at  (3,0) {\small$R_y$};
        \node[state] (rw) at  (-3,0) {\small$R_w$};

		\path[->] (X) edge node {} (Z);
        \path[->] (Z) edge node {} (Y);
        \path[->] (X) edge node {} (W);
        \path[->] (Y) edge node {} (W);
        \path[->] (W) edge node {} (ry);
        
    \end{tikzpicture}

%% file: tikz/self_missing1.tex
\begin{tikzpicture}[
		scale=0.4,
        > = stealth, 
        shorten > = 1pt, 
        auto,
        node distance = 3cm, 
        semithick 
    ]

    \tikzstyle{every state}=[
    	circle,
        draw = black,
        thick,
        minimum size = 8mm
    ]

    \node[state,fill = gray] (X) at  (0,0){$X$};
    \node[state, fill = white] (Y) at  (3,0) {$Y$};
    \node[state] (rx) at  (0,-3) {$R_x$};
    \path[->] (X) edge node {} (rx);
    
\end{tikzpicture}

%% file: method.tex
\section{Proposed method: Missing-value PC}
\label{sec:method}
In this section, we present our proposed approach, Missing-Value PC (MVPC), for causal discovery in the presence of missing data based on PC. We introduce the general MVPC framework in Section \ref{sec:mvpc_overview}, the detection of the direct causes of missingness in Section \ref{sec:detection}, and our correction methods for removing extraneous edges in Section \ref{sec:corr}. 

\subsection{Overview of MVPC}
\label{sec:mvpc_overview}
Algorithm \ref{alg:mvpc} summarizes the framework of MVPC. We perform TD-PC on $\mathbf{V}$ (Step \ref{stp:deletPC}), and then properly include $\mathbf{R}$ in the graph (Step \ref{stp:det-cas}). This is equivalent to performing TD-PC on $\mathbf{V} \cup \mathbf{R}$ under Assumption \ref{ass:re0}, \ref{ass:re1}, and \ref{ass:re2}. We then identify potential extraneous edges (Step \ref{stp:iden}). These are the edges between variables of which direct common effects are missingness indicators or ancestors of missingness indicators. Since we do not have orientation information at this stage, we cannot directly locate such extra edges; however, we can find potentially incorrect edges, as a superset of the incorrect edges. Next, we perform correction for these candidate edges (Step \ref{stp:cor}). Finally, we orient edges of the recovered causal skeleton with the same procedure as the PC algorithm.

\begin{algorithm}
\caption{Missing-value PC}\label{alg:mvpc}
\begin{algorithmic}[1]
\State \textit{Skeleton search with test-wise deletion PC}: \label{stp:deletPC}
\begin{enumerate}[label={\alph*}]
\item \textit{Graph initialization}: Build a complete undirected graph $G$ on the node set $\mathbf{V}$. 
\item \textit{Causal skeleton discovery}: Remove edges in $G$ with the same procedure as the PC algorithm \citep{Spirtes01} with the test-wise deleted data.
\end{enumerate}

\State \textit{Detecting direct causes of missingness indicators}:\label{stp:det-cas}
\Statex For each variable $V_i \in \mathbf{V}$ containing missing values and for each $j$ that $j \neq i$, test the CI relation of $R_i$ and $V_j$. If they are independent given a subset of $\mathbf{V} \setminus \{V_i, V_j\}$, $V_j$ is not a direct cause of $R_i$. 

\State \textit{Detecting potential extraneous edges}: \label{stp:iden}
\Statex For each $i\neq j$, if $V_i$ and $V_j$ are adjacent and have at least one common adjacent variable or missingness indicator, the edge between $V_i$ and $V_j$ is potentially extraneous.

\State \textit{Recovering the true causal skeleton}: 
\Statex Perform correction methods for removing the extraneous edges in $G$ as shown in Section \ref{sec:corr}.
\label{stp:cor}

\State \textit{Determining the orientation}:
\Statex Orient edges in $G$ with the same orientation procedure as the PC algorithm.
\end{algorithmic}
\end{algorithm}

\subsection{Detection of the direct causes of missingness indicators}\label{sec:detection}
In Step \ref{stp:det-cas} of Algorithm \ref{alg:mvpc}\hk{,} detecting direct causes of missingness indicators is implemented by the causal skeleton discovery procedure of TD-PC. For each missingness indicator $R_i$\hk{,} the causal skeleton discovery procedure checks all the CI relations between $R_i$ and variables in $\mathbf{V}\setminus V_i$, and then tests whether $R_i$ is conditionally independent of a variable $V_j \in \mathbf{V}\setminus V_i$ given any variable or set of variables connected to $R_i$ or $V_j$. If they are conditionally independent,  the edge between $R_i$ and $V_j$ is removed. Under Assumptions \ref{ass:re0}$\sim$\ref{ass:re2}, no extraneous edge is produced in this step because according to Proposition \ref{prop:iden}, an extraneous edge only occurs when $R_i$ and $V_j$ have at least one direct common effect. Since we assume that the missingness indicator has no child variable in Assumption \ref{ass:re0}, $R_i$ has no child, and there is no direct common effect of $R_i$ and $V_j$. Therefore, in the result of this step all the variables adjacent to $R_i$ are its direct causes.

\subsection{Recovery of the true causal skeleton} \label{sec:corr}
As shown in Section \ref{sec:problem_setting}, TD-PC produces extraneous edges in the causal skeleton, resulting in the situations of Proposition \ref{prop:iden}. In this section, we introduce our correction methods to remove the extraneous edges. We first introduce Permutation-based Correction (PermC) with an example. We then show that PermC handles most of the missingness cases. Next, we propose an alternative solution, named Density Ratio Weighted correction (DRW), for the cases which PermC does not cover.

\subsubsection{Permutation-based correction}
\paragraph{PermC in continuous cases.}
We use an example in continuous cases to demonstrate how to remove the extraneous edges with PermC. For example, suppose that we have a dataset with missing values of which the underlying missingness graph is shown in Figure \ref{fig:mar}. As discussed in Section \ref{sec:problem_setting}, when applying TD-PC to this dataset, we produce an extraneous edge between $X$ and $Y$ in the output of TD-PC. The problem is that data samples from joint distribution $P(X,Y,Z)$ are not available in the observed dataset. In this case, we test the CI relations in the test-wise deleted data from $P(X,Y^*,Z \mid R_y=0)$, producing the extraneous edge. 

PermC solves this problem by testing the CI relations in the reconstructed virtual dataset utilizing the observed data concerning 
\begin{align} \label{eq:eg1}
P(X,Y,Z) &=\int_{W} P(X,Y,Z\mid W) 
P(W) dW \nonumber \\ 
&= \int_{W} P(X,Y^*,Z\mid W,R_y=0) 
P(W) dW,
\end{align}
such that reconstructed data follow the joint distribution $P(X,Y,Z)$. 
As shown in the first step of Equation \ref{eq:eg1}, we introduce a random variable $W$ which is the direct cause of $R_y$ in Figure \ref{fig:mar} to reconstruct the dataset and then marginalize it out. With $W$, the joint distribution $P(X, Y, Z)$ is estimated by 1) learning the model for $P(X,Y,Z \mid W)$ from test-wise deleted data, 2) plugging in the values of $W$ in the dataset, as data samples from $P(W)$, and 3) disregarding the input $W$ and keeping the generated virtual data for $\{X,Y,Z\}$ to marginalize $W$ out. Given virtual data of $X$, $Y$, and $Z$ that follow the joint distribution $P(X,Y,Z)$, one can test CI relations in the complete data.

Now the issue is that the data samples from $P(X,Y,Z \mid W)$ are not directly available.
Nevertheless, we learn a model for $P(X,Y^*,Z\mid W,R_y = 0)$ to generate  virtual data of $X$, $Y$, and $Z$ from $W$, as shown in the second step of Equation \ref{eq:eg1}. Under Assumptions \ref{ass:re0}$\sim$\ref{ass:re2} we have $P(X, Y,Z\mid W)=P(X,Y^*,Z\mid W,R_y = 0)$ because $R_y \dsep \{X, Y, Z\}\mid W$; moreover, data samples from $P(X,Y^*,Z\mid W,R_y = 0)$ can be constructed by test-wise deletion. For simplicity, under the linear Gaussian assumption we apply linear regression to learning the model for $P(X,Y^*,Z\mid W, R_y=0)$ as :
\begin{equation} \label{eq:reg}
X = \alpha_{1}W  + \epsilon_{1},~~~
Y = \alpha_{2} W  + \epsilon_{2},~~~
Z = \alpha_{3} W  + \epsilon_{3},
\end{equation}
where $\alpha_{i}$ is the parameter of linear regression models and $\epsilon_{i}$ is the residual. 

Next, we sample the input values from the probability distribution $P(W)$. Estimating $P(W)$ for sampling input values is unnecessary in this case because we have the complete data of $W$, which follow $P(W)$. However, to generate virtual data with linear regression models, we cannot directly input the test-wise deleted data of $W$ and add the residuals from the linear regression models in Equation \ref{eq:reg}. In this way, the input values follow the conditional distribution $P(W|R_y=0)$ instead of $P(W)$. Thus, we shuffle the values of $W$ in the observed dataset such that $P(W^S\mid R_y=0)=P(W^S)$ where $W^S$ denotes the shuffled $W$. We then feed test-wise deleted values of $W^S$ into the linear regression models as :
\begin{equation} \label{eq:pred}
\widehat{X} :=  \alpha_{1} W^S  + \epsilon_{1},~~~
\widehat{Y} :=   \alpha_{2} W^S + \epsilon_{2},~~~
\widehat{Z} :=   \alpha_{3} W^S  + \epsilon_{3},
\end{equation}
where we denote the random variables with generated virtual values by $\widehat{X}$, $\widehat{Y}$, and $\widehat{Z}$. Finally, we test for the CI relation among $\widehat{X}$, $\widehat{Y}$, and $\widehat{Z}$.
PermC for this example is summarized in Algorithm \ref{alg:gen-var}.
\begin{algorithm}[h]
\caption{Permutation-based correction}\label{alg:gen-var}
\textbf{Input:} data of the concerned variables, such as $X$, $Y$, and $Z$ in Figure \ref{fig:mar}, and the direct causes of their corresponding missingness indicators, such as the direct cause $W$ of $R_y$ in Figure \ref{fig:mar}.\\
\textbf{Output:} The CI relations among concerned variables, such as the CI relations among $X$, $Y$, and $Z$.
\begin{algorithmic}[1]
\State Delete records containing any missing value. We denote the deleted dataset by $D_d$, and denote the original dataset by $D_o$. 
\State Regress $X$, $Y$, and $Z$ on $W$ with $D_d$ as Equation \ref{eq:reg}.\label{step:lm}
\State Shuffle data of $W$ in $D_o$, denoted by ${W}^S$, and delete records containing any missing value in $D_o$ (included $W^S$). \label{step:sf}
\State Generate virtual data of $\widehat{X}$, $\widehat{Y}$, and $\widehat{Z}$, with  $W^S$ and the residuals according to Equation \ref{eq:pred}.
\label{step:pred}
\State Test the CI relations among $\widehat{X}$, $\widehat{Y}$, and  $\widehat{Z}$ in the generated virtual data. \label{step:test}\\
\Return The CI relations among $X$, $Y$, and $Z$.
\end{algorithmic}
\end{algorithm}

Without loss of generality, we summarize the conditions under which PermC correctly removes extraneous edges. 
Suppose that we need to test the CI relation of $X$ and $Y$ given $\mathbf{Z}\subseteq \mathbf{V} \setminus \{X,Y\}$ in the generated virtual data. We denote the direct causes of missingness indicators by $Pa(\mathbf{R})$. 
The conditions for the validity of PermC are as follows.
\begin{enumerate}[label={(\roman*)}]
\item \label{cond:corr1} $\{R_x,R_y,\mathbf{R_z},\mathbf{R_w}\} \dsep \{X,Y,\mathbf{Z}\}\mid \mathbf{W}$, where the variable set $\mathbf{W}$ is the set of direct causes of missingness indicators $R_x$, $R_y$, and $\mathbf{R_z}$;
if variables in $\mathbf{W}$ also have missing values, the direct causes of their missingness indicators $ \mathbf{R_w}$ are also included in $ \mathbf{W}$;
formally, $ \mathbf{W} = Pa(R_x,R_y,\mathbf{R_z},\mathbf{R_w})$;
\item \label{cond:corr2} In the missingness graph, the missingness indicators of $\mathbf{W}$ follow the condition that $X \dsep Y \mid \mathbf{Z} \Longleftrightarrow X \dsep Y \mid \{\mathbf{Z}, \mathbf{R_w}\}$.
\end{enumerate}
Under Conditions \ref{cond:corr1} and  \ref{cond:corr2}, we have 
\begin{multline}\label{eq:cor1}
P(X,Y,\mathbf{Z}\mid \mathbf{R_w} = \mathbf{0}) = \int_\mathbf{W^*} P(X^*,Y^*,\mathbf{Z^*}\mid \mathbf{W^*},R_x=0,R_y=0,\mathbf{R_z}=\mathbf{0},\mathbf{R_w}=\mathbf{0}) \times \\
P( \mathbf{W^*}\mid \mathbf{R_w}= \mathbf{0}) d\mathbf{W^*}.
\end{multline}

To test the CI relation of $X$ and $Y$ given $\mathbf{Z}$ in data samples from $P(X,Y,\mathbf{Z})$, it is valid to test the CI relation in the generated data samples from $P(X,Y, \mathbf{Z}\mid \mathbf{R_w}= \mathbf{0})$. Under  Condition \ref{cond:corr2} the conditional independence/dependence relations in $P(X,Y, \mathbf{Z})$ also hold in  $P(X,Y, \mathbf{Z}\mid \mathbf{R_w}= \mathbf{0})$. Moreover, linear regression models in PermC are valid. Under Condition \ref{cond:corr1}, we have $P(X,Y, \mathbf{Z}\mid \mathbf{ W}, R_x=0,R_y=0, \mathbf{R_z}= \mathbf{0},  \mathbf{R_w}= \mathbf{0})=P(X,Y, \mathbf{Z}\mid  \mathbf{W})$, in which $X$, $Y$, and $\mathbf{Z}$ are conditionally Gaussian distributed given $\mathbf{W}$. Thus, we use linear regression to estimate $P(X^*,Y^*, \mathbf{Z^*}\mid  \mathbf{W^*}, R_x=0,R_y=0, \mathbf{R_z}= \mathbf{0},  \mathbf{R_w}= \mathbf{0})$ and use them in the correction.

\begin{algorithm}[H]
\caption{Binary Permutation-based correction}\label{alg:Bgen-var}
\textbf{Input:} data of the concerned variables, such as $X$, $Y$, and $Z$ in Figure \ref{fig:mar}, and the direct causes of their corresponding missingness indicators, such as the direct cause $W$ of $R_y$ in Figure \ref{fig:mar}.\\
\textbf{Output:} The CI relations among concerned variables, such as the CI relations among $X$, $Y$, and $Z$.
\begin{algorithmic}[1]
\State Delete records containing any missing value. We denote the deleted data set by $D^d$, and denote the original data set by $D^o$. 
\State Separate $D^d$ into $D_{W=0}^d$ and $D_{W=1}^d$ according to the values of $W$, e.g., $D_{W=0}^d$ are the samples in $D^d$ of which the value of $W$ is $0$.
\State Estimate the joint distributions of $X$, $Y$, and $Z$ with $D_{W=0}^d$ and $D_{W=1}^d$, denoted by $P_{W=0}(X,Y,Z)$ and $P_{W=1}(X,Y,Z)$.
\State Shuffle data of $W$ in $D^o$, denoted by ${W}^S$, and delete records containing any missing value in $D^o$ (included $W^S$). We denote this data set by $D^d_S$.  \label{step:sf}
\State Generate virtual data by sampling from $P_{W=0}(X,Y,Z)$ or  $P_{W=1}(X,Y,Z)$ depending on the values of $W^S$ in $D^d_S$.
\label{step:pred}
\State Test the CI relations among $\widehat{X}$, $\widehat{Y}$, and  $\widehat{Z}$ in the generated virtual data. \label{step:test}\\
\Return The test results of CI relations among $\widehat{X}$, $\widehat{Y}$, and  $\widehat{Z}$.
\end{algorithmic}
\end{algorithm}

\paragraph{PermC in binary cases.}
In the binary case, PermC follows the same procedures as shown in Algorithm \ref{alg:gen-var} to correct the extraneous edges. The only difference between the continuous case and the binary case is how to generate the virtual data. Taking the same missingness graph in Figure \ref{fig:mar} as an example, we generate the virtual data of which the distribution is $P(X, Y, Z)$ to test the CI relations of $X$, $Y$, and $Z$. The virtual data distribution is 
\begin{align} \label{eq:eg5}
P(X,Y,Z) &=\sum_{W} P(X,Y,Z\mid W) 
P(W) 
= \sum_{W} P(X,Y^*,Z\mid W,R_y=0) 
P(W).
\end{align}
Instead of using linear regression to generate the virtual data, we directly estimate the conditional distribution  $P(X,Y^*,Z\mid W,R_y=0)$ from the test-wise deleted data and generate data of $X$, $Y$, and $Z$ based on values of $W$. Meanwhile, we keep the marginal distribution of $W$. PermC in the binary case is summarized in Algorithm \ref{alg:Bgen-var}. Note that Condition \ref{cond:corr1} of PermC in binary cases is that 
$R_i \dsep V_i \mid \mathbf{W} $, where $i \in \{x,y,\mathbf{z}\}$ and $X$ denoted by $V_x$ here; and $R_{w_j} \dsep W_j \mid \mathbf{W}_k$, where $\mathbf{W}_k \subset \mathbf{W} \setminus {W_j}$. We denote this condition as Condition \ref{cond:corr1}$^*$. Condition \ref{cond:corr1}$^*$ is weaker than Condition \ref{cond:corr1} because there is no linear Gaussian constraint for modeling $P(X,\,Y^*\,,Z\mid W, R_y = 0)$.

\subsubsection{Density ratio weighted correction} 
\paragraph{DRW in continuous cases.} 
DRW removes  extraneous edges in situations where Condition \ref{cond:corr1} and Condition \ref{cond:corr2} are not satisfied (e.g., Figure \ref{fig:mnar}).
In these cases, we consistently estimate the joint distribution $P(\mathbf{V_a})$ of concerned variables in a CI test according to Theorem 2 of \citep{mohan2013graphical}.
Here, $\mathbf{R}$ represents the missingness indicators of $\mathbf{V_a}$. Equation \ref{eq:corr2} provides a way to reconstruct the observed dataset:
\begin{align}\label{eq:corr2}
    P(\mathbf{V_a}) &= \frac{P(\mathbf{R}=\mathbf{0}, \mathbf{V_a})}{\prod_i P(R_i=0\mid Pa(R_i),R_{Pa(R_i)}=0)}\nonumber \\  
    & = P(\mathbf{V_a}\mid \mathbf{R}=\mathbf{0}) \times c \times \prod_i \beta_{Pa(R_i)}\,,
\end{align} where $c = \frac{P(\mathbf{R}=\mathbf{0})}{\prod_i P(R_i=0\mid R_{Pa(R_i)}=0)}$ and $\beta_{Pa(R_i)} = \frac{P(Pa(R_i)\mid R_{Pa(R_i)}=0)}{P(Pa(R_i) \mid R_i=0, R_{Pa(R_i)}=0)}$.
In the second line of Equation \ref{eq:corr2}, every (conditional) probability distribution can be consistently estimated. We first apply test-wise deletion to the observed data of $\mathbf{V_a}$. 
Then, we reweight such data with the density ratios $\prod_i \beta_{Pa(R_i)}$ and the normalizing constant $c$. We estimate density ratios $\prod_i \beta_{Pa(R_i)}$ with the kernel density estimation \citep{sheather1991reliable} and compute the normalizing constant $c$. Finally, we test CI relations of concerned variables in the reweighted data samples from their corresponding joint distribution.

\paragraph{DRW in binary cases.} DRW removes extraneous edges by testing CI relations in the full data distribution, such as $P(\mathbf{V_a})$ in Equation \ref{eq:corr2}. In the binary case, the full data distribution $P(\mathbf{V_a})$ can be simplified by Equation \ref{eq:corr3} 
\begin{align}\label{eq:corr3}
    P(\mathbf{V_a}) &=   P(\mathbf{V_a}\mid \mathbf{R}=\mathbf{0}) \times \prod_i \beta_{Pa(R_i)} \times P(\mathbf{R}=\mathbf{0}),
\end{align}
where $\beta_{Pa(R_i)} = \frac{1}{ P(R_i=0\mid Pa(R_i),R_{Pa(R_i)}=0)}$ .
\footnote{Note that every term in Equation \ref{eq:corr3} can be consistently estimated by counting the number of different variable-value combinations, e.g., $\mathbf{V_a}=\mathbf{v_a}$, $\mathbf{R} = \mathbf{0}$, $Pa(R_i)=0$, and $R_{Pa(R_i)}=0$, in the observed data.} 
In Equation \ref{eq:corr3}, $P(\mathbf{V_a}\mid \mathbf{R}=\mathbf{0})$ can be regarded as the distribution of the test-wise deleted data and $\prod \beta_{Pa(R_i)} P(\mathbf{R}=\mathbf{0})$ can be considered as the weights of each data point. The test-wise deleted data with the weights are used for the weighted $G^2$ test. Finally, based on the conditional independence results of weighted $G^2$ tests, we remove extraneous edges.

%% file: experiments.tex
\section{Experiments}
\label{sec:exp}
We evaluate our method, MVPC, on synthetic data, simulated neuropathic pain diagnosis records, and real-world datasets. We first show experimental results on synthetic data (Section \ref{sec:synthetic}). We then apply the neuropathic pain diagnosis simulator to generating data in the presence of MNAR to compare different methods (Section \ref{sec:exp_sim}). Finally, we apply our method to two healthcare datasets where data entries are significantly missing. The first (Section \ref{sec:CogUSA}) is from the Cognition and aging USA (CogUSA) study \citep{mcardle2015cognition}, and the second (Section \ref{sec:atr}) is about Achilles Tendon Rupture (ATR) rehabilitation research study \citep{praxitelous2017microcirculation, domeij2016ageing}.
MVPC demonstrates superior performance compared to multiple baseline methods. 

\subsection{Synthetic data evaluation}
\label{sec:synthetic}
To best demonstrate the behavior of different causal discovery methods, we first perform the evaluation on synthetic data sampled from randomly generated causal graphs. In this section we start from the evaluation of baseline methods for continuous variable cases with the consideration of the number of samples, the number of the parents of missingness indicators, different MNAR generation methods, and the number of substantive variables (that are the variables except proxy variables and missingness indicators in a missingness graph). Since the performance of baseline methods for binary variable cases is similar to that of continuous variables, we only show the results of binary variables in large sample size. We use Structural Hamming Distance (SHD) \citep{tsamardinos2006max} and the F1 score as the evaluation metrics. Lower value of SHD is better, and higher value of F1 score is better. 

We apply both permutation-based and density ratio weighted correction method to MVPC, denoted by MVPC-PermC and MVPC-DRW. We also use test-wise deletion PC algorithm (TD-PC) as one of the baseline methods. Additionally, we apply the PC algorithm to the oracle data (without missing data), denoted by "ideal". Moreover, to decouple the effect of sample size, we construct datasets that are MCAR which have the same sample size as the average sample size of all CI tests in TD-PC. We denote PC with such virtual MCAR data by "target". 

\begin{figure}[!ht]%
\centering
\captionsetup[subfigure]{aboveskip=4pt}
\begin{subfigure}[t]{.48\textwidth}
  \centering
  \scalebox{0.65}{\input{tikz/conti_asymp_mar.tex} }
  \caption{ Structural Hamming Distance (MAR).}
  \label{fig:mar_shd_nos}
\end{subfigure}
\hspace{0.8em}%
\captionsetup[subfigure]{aboveskip=4pt}
\begin{subfigure}[t]{.48\textwidth}
  \centering
 \scalebox{0.65}{ \input{tikz/conti_asymp_mar_f1.tex}  }
  \caption{F$1$ scores (MAR).}
  \label{fig:mar_f1_nos}
\end{subfigure}
\hspace{0.8em}%
\captionsetup[subfigure]{aboveskip=4pt}

\begin{subfigure}[t]{.48\textwidth}
  \centering
  \scalebox{0.65}{\input{tikz/conti_asymp_mnar.tex} }
  \caption{ Structural Hamming Distance (MNAR).}
  \label{fig:mnar_shd_nos}
\end{subfigure}
\hspace{0.8em}%
\captionsetup[subfigure]{aboveskip=4pt}
\begin{subfigure}[t]{.48\textwidth}
  \centering
 \scalebox{0.65}{ \input{tikz/conti_asymp_mnar_f1.tex}  }
  \caption{F$1$ scores (MNAR).}
  \label{fig:mnar_f1_nos}
\end{subfigure}
\hspace{0.8em}%
\captionsetup[subfigure]{aboveskip=4pt}
\caption{ Results of the baseline methods in continuous cases with different sample sizes.
The methods are evaluated with Structural Hamming Distance (SHD) and F$1$ score. Lower value of SHD is better. Higher value of F$1$ score is better. The data are Missing At Random (MAR) in Panal (a) and (b); Missing Not At Random (MNAR) in Panal (c) and (d).}
\label{fig:mar-mnar-nos}
\end{figure}
\paragraph{Number of samples.}
We first study the performance of baseline methods with different number of samples, following the procedures in \citep{colombo2012learning, strobl2017fast} which randomly generates Gaussian DAG and samples data based on the given DAG. The results are shown in Figure \ref{fig:mar-mnar-nos}. We generate synthetic datasets that are MAR and MNAR with the number of sample size $500$, $1000$, $5000$, $10000$, $50000$, and $100000$. The number of substantive variables is $20$ in every causal graph. 
In the datasets that are MAR, $10$ variables have missing values and at most $5$ missingness indicators of them are in the case of Proposition \ref{prop:iden}. Moreover, the parents of these $10$ missingness indicators have no missing values. In this experiment we limit the number of the parents of missingness indicators to $1$. In the following experiments we will study the influence of the number of such parents. 

For generating the datasets that are MNAR, we firstly select at most $5$ missingness indicators that are in the case of Proposition \ref{prop:iden}, and then randomly assign missing values for the parents of such missingness indicators. 
We call such a way, which generates MNAR datasets in a 2-step manner, "MNAR: MAR + MCAR". 
There is another more general way named "MNAR: no self-masking". This MNAR generation method first selects at most $5$ missingness indicators that are in the case of Proposition \ref{prop:iden}. Then, it assigns missing values for each direct cause of missingness indicators selected in the first step, considering the values of another variable which cannot be the direct cause itself. In this experiment we only generate MNAR datasets with "MNAR: MAR + MCAR".

Figure \ref{fig:mar-mnar-nos} shows that our proposed algorithm, MVPC, consistently has superior performance compared to TD-PC, and is converging to the "target" performance with increasing the sample size across both metrics. 
Moreover, we find that MVPC-PermC is more data-efficient than MVPC-DRW regarding the convergence speed of both methods. 
More interestingly, TD-PC is diverging and might be asymptotically wrong, which could depend on how to generate missing values.

\begin{figure}[!ht]%
\centering
\captionsetup[subfigure]{aboveskip=4pt}
\begin{subfigure}[t]{.42\textwidth}
  \centering
  \scalebox{0.65}{\input{tikz/conti_mul_mar.tex} }
  \caption{ Structural Hamming Distance.}
  \label{fig:nop_shd}
\end{subfigure}
\hspace{0.8em}%
\captionsetup[subfigure]{aboveskip=4pt}
\begin{subfigure}[t]{.42\textwidth}
  \centering
 \scalebox{0.65}{ \input{tikz/conti_mul_mar_f1.tex}  }
  \caption{F$1$ scores.}
  \label{fig:nop_f1}
\end{subfigure}
\hspace{0.8em}%
\captionsetup[subfigure]{aboveskip=4pt}
\caption{
Results of the baseline methods for the study: the influence of the parent number of missingness indicators.}
\label{fig:mar-nop}
\end{figure}
\paragraph{Number of the parents of missingness indicators.} 
We also investigate the influence of the parent number of missingness indicators. In this experiments, we remove the limitation of the parent number of missingness indicators, and generate MAR datasets. For the purpose of this experiment it is not necessary to use MNAR which introduces much complexity. The number of substantive variables are $20$. The number of missingness indicators are $10$, and at most $5$ missingness indicators are in the case of Proposition \ref{prop:iden}. The sample size is $50000$. 

Figure \ref{fig:mar-nop} displays that most of the methods have no significant difference between the experiments of single parent and multiple parents, and that MVPC-DRW is sensitive to the number of parents of missingness indicators. The reason could be that the density ratio estimation is implemented by the density estimation method which performs well for low-dimension data (of which the dimension is lower than $4$). When a missingness indicator can have multiple causes, MVPC is more likely to estimate the probability distribution containing more than $3$ variables, which leads to its worse performance than that in the single parent experiments.

\paragraph{Comparison of MNAR generation methods.} 
We compare the performance of baseline methods in the MNAR datasets with different generation methods: "MNAR: no self-masking" and "MNAR: MAR + MCAR". The sample size is $100000$, and the number for substantive variables is $20$. The missingness indicators is $10$, and at most $5$ missingness indicators are in the case of Proposition \ref{prop:iden}.
\begin{figure}[!ht]%
\centering
\captionsetup[subfigure]{aboveskip=4pt}
\begin{subfigure}[t]{.42\textwidth}
  \centering
  \scalebox{0.65}{\input{tikz/mnar_mech.tex} }
  \caption{ Structural Hamming Distance.}
  \label{fig:mnar-g-shd}
\end{subfigure}
\hspace{0.8em}%
\captionsetup[subfigure]{aboveskip=4pt}
\begin{subfigure}[t]{.42\textwidth}
  \centering
 \scalebox{0.65}{ \input{tikz/mnar_mech_f1.tex}  }
  \caption{F$1$ score.}
  \label{fig:mnar-g-f1}
\end{subfigure}
\hspace{0.8em}%
\captionsetup[subfigure]{aboveskip=4pt}
\caption{
Results of the baseline methods in the datasets with different MNAR generation methods.}
\label{fig:mnar-g}
\end{figure}

Figure \ref{fig:mnar-g} shows that most of the methods perform better in the datasets that are generated with "MNAR: MAR + MCAR". We also find that although MVPC-Permc performs better than MVPC-DRW, the average performance of MVPC-PermC in different datasets varies more than that of MVPC-DRW. 
The results are as expected because the missingness graphs generated by "MNAR: MAR + MCAR" satisfy the conditions of PermC correction method; however, the the MNAR missingness graphs generated by "MNAR: no self-masking" cannot.
Even though the MNAR missingness graphs generated by "MNAR: no self-masking" cannot satisfy the conditions of "MNAR-PermC", the result of MVPC-PermC is still close to the "target". This also indicates that MVPC-PermC can handle most of the MNAR cases.

\paragraph{Number of substantive variables.}
In this experiment we study the performance of baseline methods with different number of substantive variables. As Figure \ref{fig:mnar-nov} shown, we generate three groups of missingness graphs with "MNAR: MAR + MCAR" containing $20$, $50$, $100$ substantive variables. Respectively, they have $10$, $16$, and $20$ missingness indicators; and at most $5$, $8$, and $10$ missingness indicators that are in the case of Proposition \ref{prop:iden}. The sample size is $10000$. Figure \ref{fig:mnar-nov} displays that MVPC performs well in the large number of variables and better than Test-wise deletion PC that is severely influenced by the number of substantive variables. 
\begin{figure}[!ht]%
\centering
\captionsetup[subfigure]{aboveskip=4pt}
\begin{subfigure}[t]{.42\textwidth}
  \centering
  \scalebox{0.65}{\input{tikz/mnar_NoV.tex} }
  \caption{ Structural Hamming Distance.}
  \label{fig:nov_shd}
\end{subfigure}
\hspace{0.8em}%
\captionsetup[subfigure]{aboveskip=4pt}
\begin{subfigure}[t]{.42\textwidth}
  \centering
 \scalebox{0.65}{ \input{tikz/mnar_NoV_f1.tex}  }
  \caption{F$1$ scores.}
  \label{fig:nov_f1}
\end{subfigure}
\hspace{0.8em}%
\captionsetup[subfigure]{aboveskip=4pt}
\caption{
Results of the baseline methods in MNAR datasets with different number of substantive variables.}
\label{fig:mnar-nov}
\end{figure}

\paragraph{Binary data evaluation.}
To evaluate MVPC in the binary case, we used Tetrad \citep{spirtes2004tetrad} to generate binary datasets together with corresponding ground-truth causal graphs. 
We compare the performance of baseline methods in MAR and MNAR datasets. The sample size is $600000$. The number of substantive variables is $20$ in each dataset. The number of missingness indicators is $10$ and at most $5$ of them are in the case of Proposition \ref{prop:iden}. We limit the number of the parent of missingness indicators to $1$. We use "MNAR: MAR + MCAR" to generate MNAR data. The values of missingness indicators follow the Bernoulli distribution of which the parameters depend on the parent value of missingness indicators.

Figure \ref{fig:bin_mar-mnar} shows that the results of MVPC are clearly better than TD-PC and close to "target" in the result with SHD metric. Moreover, we find that MVPC-DRW in binary cases is much less data-efficiency than MVPC-PermC in continuous cases due to the fact that MVPC-DRW starts performing better than TD-PC when the samples size is larger than $500000$.

\begin{figure}[!ht]%
\centering
\captionsetup[subfigure]{aboveskip=4pt}
\begin{subfigure}[t]{.48\textwidth}
  \centering
  \scalebox{0.65}{\input{tikz/bin_mnar_mar.tex} }
  \caption{ Structural Hamming Distance.}
  \label{fig:bin_mar_shd}
\end{subfigure}
\hspace{0.8em}%
\captionsetup[subfigure]{aboveskip=4pt}
\begin{subfigure}[t]{.48\textwidth}
  \centering
 \scalebox{0.65}{ \input{tikz/bin_mnar_mar_f1.tex}  }
  \caption{F$1$ score.}
  \label{fig:bin_mnar_shd}
\end{subfigure}
\hspace{0.8em}%
\captionsetup[subfigure]{aboveskip=4pt}
\caption{ Results of the baseline methods in binary cases that are Missing At Random (MAR) and Missing Not At Random (MNAR). }
\label{fig:bin_mar-mnar}
\end{figure}

\subsection{Neuropathic Pain Diagnosis Simulation data evaluation}\label{sec:exp_sim}
The evaluation of causal discovery methods is lack of real-world benchmark data sets, which has been a challenge \citep{pannel2018}. 
We commonly evaluate on synthetic data sampled from randomly generated causal graphs (e.g., the synthetic data in Section \ref{sec:synthetic}). However, synthetic data are mostly based on the model proposed in a work such that it could be inappropriate to evaluate others' works. Moreover, the synthetic data experiments may show the superior performance of proposed methods but sometimes may oversimplify the challenges in real-world scenarios \citep{garant2016evaluating}. Fortunately, towards tackling the evaluation challenge,  well-understood causal influences are available in some specific scenarios of disciplines such as biology and physics. This gives us opportunities to utilize domain knowledge and build realistic simulators. In this way, we can generate data from simulators and use them as benchmark datasets for the evaluation of causal discovery algorithms.

We develop a simulator to generate neuropathic pain diagnostic records\footnote{ The simulator is available at \href{https://github.com/TURuibo/Neuropathic-Pain-Diagnosis-Simulator}{https://github.com/TURuibo/Neuropathic-Pain-Diagnosis-Simulator}.}. The neuropathic pain diagnostic simulator \citep{tu2019neuropathic} is an attempt of evaluating causal discovery methods which fills the the gap between the randomly generated dataset and the real-world dataset (e.g., the data sets in Section \ref{sec:CogUSA} and Section \ref{sec:atr}). The simulator considers each diagnostic label as a variable of which the value $1$ indicates that the diagnostic label exists in a patient record and $0$ otherwise. 
In total, $222$ binary variables (nodes in a causal graph) and $770$ pairs of causal relations (edges in a causal graph) are included in a causal graph considering the experiences and the domain knowledge of clinicians. The causal graph contains different d-separations such as the folk structure, the collider structure, and the chain structure. The complete interactive causal graph is visualized at: \url{https://cutt.ly/BekNFSy}. Other details about the simulator can be found in \citep{tu2019neuropathic}.

\begin{table}[H]
\center
\caption{ Results of applying causal discovery methods to  simulation data with missing values from the Neuropathic Pain diagnosis simulator \citep{tu2019neuropathic}.}
\begin{tabular}{lllll}
\hline
         & Cau\_acc & Recall & Precision & F1-score \\ \hline
PC-ideal  & 0.047    &0.046  &     0.44      &0.085\\
PC-target & 0.046    &0.046   &     0.44      &0.085 \\
MVPC     & 0.045    & 0.043  & 0.45     & 0.078    \\
TD-PC    & 0.033    & 0.025  & 0.559     & 0.047    \\ \hline
\end{tabular}
\label{exp:sim}
\end{table}

In this experiment, we use the neuropathic pain diagnosis simulator to generate data that are MAR. 
The sample size is $1000$ and the other parameters are the same with the ones in \citep{tu2019neuropathic}. We then evaluate the performance of different methods with the causal accuracy ($Cau\_acc$) \citep{claassen2012bayesian}, recall, precision, and F1 score of resulted undirected graphs on the simulated data.
We use MVPC with the density ratio weighted correction method in the binary case. Table \ref{exp:sim} shows that all methods have low recall, which indicates the difficulty of applying causal discovery to this simulation dataset. Moreover, the result of MVPC is close to \textit{PC-ideal} and \textit{PC-target} and better than TD-PC in the simulation experiments ( TD-PC, \textit{PC-ideal} and \textit{PC-target} are introduced in Section \ref{sec:synthetic}).

\subsection{The Cognition and aging USA (CogUSA) study}
\label{sec:CogUSA}
In this experiment, we aim to discovery causal relations in the CogUSA study as in \citep{strobl2017fast}. This is a typical survey based healthcare dataset with a large amount variables with missing values. In this scenario, the missingness mechanism is unknown and we could expect MCAR, MAR, and MNAR occur.
\begin{figure}[!ht]
\centering
\input{tikz/CogUSA_barplot.tex}
\caption{Performance of different methods on CogUSA study. Lower cost is  better. The cost is the count of errors comparing with known causal constrains from experts.
}
\label{fig:cog}
\end{figure}

We use the same 16 variables of interest in the CogUSA study as in \citep{strobl2017fast}. Since the missingness indicators of the 16 variables can be caused by other variables, we utilize the rest variables when applying MVPC to the dataset. We use the BIC score for CI test (likelihood ratio test with the BIC penalty as the threshold). 
Figure \ref{fig:cog} shows the performance evaluated using the known causal constraints: 1) Variables are in two groups with no inter-group causal relation; 2) there are causal relations between two pairs of variables given by the domain expertise.
Each violation of these known causal relations adds $1$ in the cost shown in Figure \ref{fig:cog}. Our proposed method obtains the best performance (lowest cost) comparing with deletion-based PC and deletion-based FCI \citep{strobl2017fast}. This demonstrates the capabilities of our method in real life applications.

\subsection{Achilles Tendon Rupture study} \label{sec:atr}

In the end, we perform causal discovery on a Achilles Tendon Rupture (ATR) study dataset  \citep{praxitelous2017microcirculation, hamesse2018simultaneous}, collected in multiple hospitals \footnote{In the ATR study experiment, only Paul Ackermann and Ruibo Tu get access to the ATR dataset.}. ATR is a type of soft tissue injury involving a long rehabilitation process. Understanding causal relations among various factors and healing outcomes is essential for practitioners. 
The list-wise deletion method is not applicable for this case because about $70\%$ of the data entries are missing, which means that very rare patients have complete data. Thus, we apply our method and TD-PC to this dataset. We ran experiments on the full dataset with more than 100 variables. Figure \ref{fig:atr} shows part of the causal graph. 

We find that age, gender, BMI (body mass index), and LSI (Limb Symmetry Index) in the causal graph given by MVPC do not affect the healing outcome measured by Foot Ankle Outcome Score (FAOS). This result is consistent with \citep{praxitelous2017microcirculation, domeij2016ageing}. To test the effectiveness of MNAR, we further introduce an auxiliary variable $S$ which is generated from two variables: Operation time ($OP_{time}$) and FAOS. This variable further causes the missingness indicator of FAOS. Figure \ref{fig:atr_opt2} and \ref{fig:atr_opt1} show the results of these variables using TD-PC and our proposed method. Our proposed MVPC is able to correctly remove the extraneous edge between Operation time and FAOS. 

\begin{figure}[H]%
\centering
\begin{subfigure}{.3\textwidth}
  \centering
  \input{tikz/ATR_BMIetc.tex}
  \caption{Consistent results}
  \label{fig:atr}
\end{subfigure}
\begin{subfigure}{.3\textwidth}
  \centering
  \input{tikz/ATR_delete}  
    \caption{Test-wise deletion PC}
  \label{fig:atr_opt2}
\end{subfigure}
\begin{subfigure}{.3\textwidth}
  \centering
  \input{tikz/ATR_ours}  
    \caption{MVPC (proposed)}
  \label{fig:atr_opt1}
\end{subfigure}
\caption{Causal discovery results in the ATR study. Experiments were run over all variables. We show only a part of the whole causal graph. Panel (a) shows the relations among five variables given by MVPC. The relations are consistent with medical studies. Panel (b) and (c) show an example where MVPC is able to correct the error of TD-PC.}
\label{fig:atr_2}
\end{figure}

%% file: tikz/conti_asymp_mar.tex
\definecolor{orange}{RGB}{253,160,161}
\begin{tikzpicture}[scale=0.85]
  \begin{axis}[
    ybar,
    legend style={
    draw=none,
    font=\Large,
    at={(0.99,1.2)},
	anchor=north,legend columns=5},
    enlargelimits=0.1,
    ylabel={SHD},
    ylabel style={font=\Large},
    xlabel={Sample size},
    xlabel style={font=\Large},
	symbolic x coords={500, 1000, 5000, 10000, 50000,100000},
    enlarge y limits=upper, ymin=0,
    xtick=data,
    bar width=7pt,
    width=1.8*\textwidth,
    axis x line*=bottom,
    axis y line*=left,
    yticklabel style = {font=\Large},
    x tick label style={font=\Large}
    ]
    \addplot +[black]
    [
    fill=yellow,
    postaction={
        pattern=crosshatch dots
    },
    pattern color=black,
    draw=black,
    error bars/.cd,
    y fixed,
    y dir=both, 
    y explicit,
    ]
    coordinates{
(500,8.4)+-=(0,4.550946)
(1000,5.4)+-=(0,3.565265)
(5000,2.2)+-=(0,2.043961)
(10000,2.3)+-=(0,1.636392)
(50000,1.6)+-=(0,2.01108)
(100000,0.6)+-=(0,1.074968)
};
 \addplot
    [ 
    fill=gray,
    postaction={
    	pattern color=white,
        pattern=crosshatch dots
    },
    error bars/.cd,
    y fixed,
    y dir=both, 
    y explicit,
    ]
    coordinates{
(500,9.9)+-=(0,5.043147)
(1000,8.1)+-=(0,4.332051)
(5000,3.6)+-=(0,3.134042)
(10000,2.3)+-=(0,2.002776)
(50000,1.4)+-=(0,1.264911)
(100000,1.3)+-=(0,1.418136)
};
 \addplot +[black]
    [ 
    fill=orange,
    postaction={
        pattern=crosshatch dots
    },
    draw=black,
    error bars/.cd,
    y fixed,
    y dir=both, 
    y explicit,
    ]
    coordinates{
(500,11.7)+-=(0,4.056545)
(1000,10.3)+-=(0,3.917199)
(5000,5.5)+-=(0,3.95109)
(10000,4.6)+-=(0,3.747592)
(50000,3)+-=(0,3.265986)
(100000,1.9)+-=(0,2.078995)
};
\addplot +[black]
    [ 
    fill = cyan,
    postaction={
    	pattern color=black,
        pattern=south west lines
    },
    error bars/.cd,
    y fixed,
    y dir=both, 
    y explicit,
    ]
    coordinates{
(500,12.2)+-=(0,4.056545)
(1000,10.7)+-=(0,3.917199)
(5000,7.4)+-=(0,3.95109)
(10000,6.6)+-=(0,3.747592)
(50000,6.1)+-=(0,3.265986)
(100000,5.6)+-=(0,2.078995)

};
\addplot +[black]
    [ 
    pattern=south west lines,
    pattern color=black,
    draw=black,
    error bars/.cd,
    y fixed,
    y dir=both, 
    y explicit,
    ]
    coordinates{
(500,12.2)+-=(0,4.779586)
(1000,11)+-=(0,3.527668)
(5000,9.7)+-=(0,3.743142)
(10000,10.2)+-=(0,4.077036)
(50000,10.8)+-=(0,3.32666)
(100000,12.5)+-=(0,3.472111)
};
 \legend{ideal $\;$, target $\;$ , MVPC-PermC $\;$, MVPC-DRW $\;$, TD-PC $\;$}
  \end{axis}
  
\end{tikzpicture}

%% file: tikz/conti_asymp_mar_f1.tex
\definecolor{orange}{RGB}{253,160,161}
\begin{tikzpicture}[scale=0.85]
  \begin{axis}[
    ybar,
    legend style={
    draw=none,
    font=\Large,
    at={(0.89,1.2)},
	anchor=north,legend columns=5},
    enlargelimits=0.1,
    ylabel={F$1$ score},
    ylabel style={font=\Large},
    xlabel={Sample size},
    xlabel style={font=\Large},
	symbolic x coords={500, 1000, 5000, 10000, 50000,100000},
    enlarge y limits=upper, ymin=0.7,
    xtick=data,
    bar width=7pt,
    width=1.8*\textwidth,
    axis x line*=bottom,
    axis y line*=left,
    yticklabel style = {font=\Large},
    x tick label style={font=\Large}
    ]
    \addplot +[black]
    [
    fill=yellow,
    postaction={
        pattern=crosshatch dots
    },
    pattern color=black,
    draw=black,
    error bars/.cd,
    y fixed,
    y dir=both, 
    y explicit,
    ]
    coordinates{
(500,0.90159798)+-=(0,0.03317655)
(1000,0.93785936)+-=(0,0.02998219)
(5000,0.97701865)+-=(0,0.03031901)
(10000,0.96605412)+-=(0,0.02641106)
(50000,0.9731521)+-=(0,0.03725419)
(100000,0.99076619)+-=(0,0.01516638)
};
 \addplot
    [ 
    fill=gray,
    postaction={
    	pattern color=white,
        pattern=crosshatch dots
    },
    error bars/.cd,
    y fixed,
    y dir=both, 
    y explicit,
    ]
    coordinates{
(500,0.85787476)+-=(0,0.03757024)
(1000,0.9123071)+-=(0,0.0333262)
(5000,0.97496044)+-=(0,0.02624966)
(10000,0.9711707)+-=(0,0.03040794)
(50000,0.97957701)+-=(0,0.02068186)
(100000,0.97818199)+-=(0,0.02464026)
};
    
 \addplot +[black]
    [ 
    fill=orange,
    postaction={
        pattern=crosshatch dots
    },
    draw=black,
    error bars/.cd,
    y fixed,
    y dir=both, 
    y explicit,
    ]
    coordinates{
(500,0.776815)+-=(0,0.03644533)
(1000,0.8382578)+-=(0,0.04356013)
(5000,0.93576002)+-=(0,0.03679431)
(10000,0.95892186)+-=(0,0.02947642)
(50000,0.9773225)+-=(0,0.02551258)
(100000,0.98176237)+-=(0,0.02177808)
};
\addplot +[black]
    [ 
    fill = cyan,
    postaction={
    	pattern color=black,
        pattern=south west lines
    },
    error bars/.cd,
    y fixed,
    y dir=both, 
    y explicit,
    ]
    coordinates
{
(500,0.77024964)+-=(0,0.03380019)
(1000,0.83370939)+-=(0,0.05198677)
(5000,0.90030897)+-=(0,0.04612013)
(10000,0.91433006)+-=(0,0.03937072)
(50000,0.92554874)+-=(0,0.03951653)
(100000,0.93374738)+-=(0,0.03409349)
};
\addplot +[black]
    [ 
    pattern=south west lines,
    pattern color=black,
    draw=black,
    error bars/.cd,
    y fixed,
    y dir=both, 
    y explicit,
    ]
    coordinates{
(500,0.77114393)+-=(0,0.04020575)
(1000,0.8289911)+-=(0,0.0478328)
(5000,0.87942003)+-=(0,0.04694922)
(10000,0.88981424)+-=(0,0.05036201)
(50000,0.86141368)+-=(0,0.03062026)
(100000,0.8580337)+-=(0,0.02397)
};
  \end{axis}
\end{tikzpicture}

%% file: tikz/conti_asymp_mnar.tex
\definecolor{orange}{RGB}{253,160,161}
\begin{tikzpicture}[scale=0.85]
  \begin{axis}[
    ybar,
    legend style={
    draw=none,
    font=\Large,
    at={(0.99,1.2)},
	anchor=north,legend columns=5},
    enlargelimits=0.1,
    ylabel={SHD},
    ylabel style={font=\Large},
    xlabel={Sample size},
    xlabel style={font=\Large},
	symbolic x coords={500, 1000, 5000, 10000, 50000,100000},
    enlarge y limits=upper, ymin=0,
    xtick=data,
    bar width=7pt,
    width=1.8*\textwidth,
    axis x line*=bottom,
    axis y line*=left,
    yticklabel style = {font=\Large},
    x tick label style={font=\Large}
    ]
    \addplot +[black]
    [
    fill=yellow,
    postaction={
        pattern=crosshatch dots
    },
    pattern color=black,
    draw=black,
    error bars/.cd,
    y fixed,
    y dir=both, 
    y explicit,
    ]
    coordinates{
(500,7.6)+-=(0,3.949684)
(1000,6)+-=(0,3.620927)
(5000,3.1)+-=(0,3.813718)
(10000,3.1)+-=(0,3.28126)
(50000,1.8)+-=(0,1.813529)
(100000,1.1)+-=(0,0.9944289)
};
 \addplot
    [ 
    fill=gray,
    postaction={
    	pattern color=white,
        pattern=crosshatch dots
    },
    error bars/.cd,
    y fixed,
    y dir=both, 
    y explicit,
    ]
    coordinates{
(500,9)+-=(0,4.570436)
(1000,7.1)+-=(0,3.3483)
(5000,3.3)+-=(0,2.790858)
(10000,3.6)+-=(0,4.060651)
(50000,1.6)+-=(0,2.01108)
(100000,1.7)+-=(0,1.828782)
};
 \addplot +[black]
    [ 
    fill=orange,
    postaction={
        pattern=crosshatch dots
    },
    draw=black,
    error bars/.cd,
    y fixed,
    y dir=both, 
    y explicit,
    ]
    coordinates{
(500,13.4)+-=(0,4.060651)
(1000,10.6)+-=(0,4.195235)
(5000,6.3)+-=(0,3.713339)
(10000,6.2)+-=(0,3.552777)
(50000,3.8)+-=(0,2.740641)
(100000,3.8)+-=(0,3.583915)
};
\addplot +[black]
    [ 
    fill = cyan,
    postaction={
    	pattern color=black,
        pattern=south west lines
    },
    error bars/.cd,
    y fixed,
    y dir=both, 
    y explicit,
    ]
    coordinates{
(500,13.6)+-=(0,4.060651)
(1000,10.7)+-=(0,4.195235)
(5000,8)+-=(0,3.713339)
(10000,9.1)+-=(0,3.552777)
(50000,8.4)+-=(0,2.740641)
(100000,7.5)+-=(0,3.583915)

};
\addplot +[black]
    [ 
    pattern=south west lines,
    pattern color=black,
    draw=black,
    error bars/.cd,
    y fixed,
    y dir=both, 
    y explicit,
    ]
    coordinates{
(500,13.3)+-=(0,4.164666)
(1000,10.6)+-=(0,3.977716)
(5000,9)+-=(0,3.887301)
(10000,10.9)+-=(0,3.725289)
(50000,11.6)+-=(0,2.221111)
(100000,11.7)+-=(0,3.198958)
};
  \end{axis}
  
\end{tikzpicture}

%% file: tikz/conti_asymp_mnar_f1.tex
\definecolor{orange}{RGB}{253,160,161}
\begin{tikzpicture}[scale=0.85]
  \begin{axis}[
    ybar,
    legend style={
    draw=none,
    font=\Large,
    at={(0.89,1.2)},
	anchor=north,legend columns=5},
    enlargelimits=0.1,
    ylabel={F$1$ score},
    ylabel style={font=\Large},
    xlabel={Sample size},
    xlabel style={font=\Large},
	symbolic x coords={500, 1000, 5000, 10000, 50000,100000},
    enlarge y limits=upper, ymin=0.65,
    xtick=data,
    bar width=7pt,
    width=1.8*\textwidth,
    axis x line*=bottom,
    axis y line*=left,
    yticklabel style = {font=\Large},
    x tick label style={font=\Large}
    ]
    \addplot +[black]
    [
    fill=yellow,
    postaction={
        pattern=crosshatch dots
    },
    pattern color=black,
    draw=black,
    error bars/.cd,
    y fixed,
    y dir=both, 
    y explicit,
    ]
    coordinates{
(500,0.90618137)+-=(0,0.04721413)
(1000,0.92596654)+-=(0,0.04660138)
(5000,0.97249609)+-=(0,0.03447395)
(10000,0.96654568)+-=(0,0.03550662)
(50000,0.96248408)+-=(0,0.03666697)
(100000,0.97470964)+-=(0,0.03135395)
};
 \addplot
    [ 
    fill=gray,
    postaction={
    	pattern color=white,
        pattern=crosshatch dots
    },
    error bars/.cd,
    y fixed,
    y dir=both, 
    y explicit,
    ]
    coordinates{
(500,0.86653942)+-=(0,0.06187828)
(1000,0.9024617)+-=(0,0.04955342)
(5000,0.96438528)+-=(0,0.02723366)
(10000,0.97271344)+-=(0,0.03190197)
(50000,0.97297679)+-=(0,0.02854369)
(100000,0.96353246)+-=(0,0.02056618)
};
    
 \addplot +[black]
    [ 
    fill=orange,
    postaction={
        pattern=crosshatch dots
    },
    draw=black,
    error bars/.cd,
    y fixed,
    y dir=both, 
    y explicit,
    ]
    coordinates{
(500,0.69378403)+-=(0,0.07805478)
(1000,0.80281703)+-=(0,0.06873515)
(5000,0.88799139)+-=(0,0.04532844)
(10000,0.92832555)+-=(0,0.05097682)
(50000,0.93195082)+-=(0,0.04196382)
(100000,0.94131148)+-=(0,0.04171016)
};
\addplot +[black]
    [ 
    fill = cyan,
    postaction={
    	pattern color=black,
        pattern=south west lines
    },
    error bars/.cd,
    y fixed,
    y dir=both, 
    y explicit,
    ]
    coordinates
{
(500,0.68937486)+-=(0,0.08418838)
(1000,0.80281703)+-=(0,0.06873515)
(5000,0.86256037)+-=(0,0.05827034)
(10000,0.87246418)+-=(0,0.03838169)
(50000,0.84438515)+-=(0,0.03728165)
(100000,0.89148505)+-=(0,0.03294932)
};
\addplot +[black]
    [ 
    pattern=south west lines,
    pattern color=black,
    draw=black,
    error bars/.cd,
    y fixed,
    y dir=both, 
    y explicit,
    ]
    coordinates{
(500,0.6979549)+-=(0,0.0739816)
(1000,0.8106143)+-=(0,0.05369446)
(5000,0.86386034)+-=(0,0.04441772)
(10000,0.8921976)+-=(0,0.0598558)
(50000,0.86012766)+-=(0,0.03229374)
(100000,0.851103)+-=(0,0.03710399)
};
  \end{axis}
\end{tikzpicture}

%% file: tikz/conti_mul_mar.tex
\definecolor{orange}{RGB}{253,160,161}
\begin{tikzpicture}[scale=0.95]
  \begin{axis}[
    ybar,
legend style={
    at={(0.25,1)},
	anchor=north,legend columns=1},
    enlargelimits=0.3,
    ylabel={SHD},
    ylabel style={font=\Large},
    xlabel style={font=\Large},
	symbolic x coords={
	$\text{ideal}$,
	$\text{target}$,
	$\text{MVPC-PermC}$,
	$\text{MVPC-DRW}$,
	$\text{TD-PC}$},
    enlarge y limits=upper, ymin=0,
    xtick=data,
    bar width=10pt,
    width=1.8*\textwidth,
    axis x line*=bottom,
    axis y line*=left,
    yticklabel style = {font=\Large},
    x tick label style={font=\Large,rotate=30,anchor=east}
    ]
    \addplot +[black]
    [
    fill=lime,
    postaction={
        pattern=crosshatch dots
    },
    pattern color=black,
    draw=black,
    error bars/.cd,
    y fixed,
    y dir=both, 
    y explicit,
    ]
    coordinates{
($\text{ideal}$,1.6)+-=(0,2.01108)
($\text{target}$, 1.5)+-=(0,1.354006)
($\text{MVPC-PermC}$,3.1)+-=(0,3.212822)
($\text{MVPC-DRW}$,8.2)+-=(0,3.212822)
($\text{TD-PC}$,9.8)+-=(0,3.084009)
};
 \addplot
    [ 
    fill=teal,
    postaction={
    	pattern color=white,
        pattern=crosshatch dots
    },
    error bars/.cd,
    y fixed,
    y dir=both, 
    y explicit,
    ]
    coordinates{
($\text{ideal}$,1.60000)+-=(0,2.01108)
($\text{target}$, 1.000000)+-=(0,1.333333)
($\text{MVPC-PermC}$,3.700000)+-=(0,3.020302)
($\text{MVPC-DRW}$,5.400000)+-=(0,3.020302)
($\text{TD-PC}$,9.600000)+-=(0,2.836273)
};
 \legend{MAR: Multiple parents, MAR: Single parent}

\end{axis}
  
\end{tikzpicture}

%% file: tikz/conti_mul_mar_f1.tex
\definecolor{orange}{RGB}{253,160,161}
\begin{tikzpicture}[scale=0.95]
  \begin{axis}[
    ybar,
legend style={
    at={(0.25,1)},
	anchor=north,legend columns=1},
    enlargelimits=0.3,
    ylabel={F1 score},
    ylabel style={font=\Large},
    xlabel style={font=\Large},
	symbolic x coords={
	$\text{ideal}$,
	$\text{target}$,
	$\text{MVPC-PermC}$,
	$\text{MVPC-DRW}$,
	$\text{TD-PC}$},
    enlarge y limits=upper, ymin=0.8,
    xtick=data,
    bar width=10pt,
    width=1.8*\textwidth,
    axis x line*=bottom,
    axis y line*=left,
    yticklabel style = {font=\Large},
    x tick label style={font=\Large,rotate=30,anchor=east}
    ]
    \addplot +[black]
    [
    fill=lime,
    postaction={
        pattern=crosshatch dots
    },
    pattern color=black,
    draw=black,
    error bars/.cd,
    y fixed,
    y dir=both, 
    y explicit,
    ]
    coordinates{
($\text{ideal}$,0.97315210)+-=(0,0.03725419)
($\text{target}$, 0.98184459)+-=(0,0.02177366)
($\text{MVPC-PermC}$,0.97179919)+-=(0,0.02853839)
($\text{MVPC-DRW}$,0.90540764)+-=(0,0.04623944)
($\text{TD-PC}$,0.89568391)+-=(0,0.05336288)
};
 \addplot
    [ 
    fill=teal,
    postaction={
    	pattern color=white,
        pattern=crosshatch dots
    },
    error bars/.cd,
    y fixed,
    y dir=both, 
    y explicit,
    ]
    coordinates{
($\text{ideal}$,0.97315210 )+-=(0,0.03725419)
($\text{target}$, 0.98620860 )+-=(0,0.02311653)
($\text{MVPC-PermC}$,0.96481106 )+-=(0,0.02816335)
($\text{MVPC-DRW}$,0.93856063 )+-=(0,0.04970662)
($\text{TD-PC}$,0.88901466 )+-=(0,0.04017103)
};
\end{axis}
  
\end{tikzpicture}

%% file: tikz/mnar_mech.tex
\definecolor{orange}{RGB}{253,160,161}
\begin{tikzpicture}[scale=0.95]
  \begin{axis}[
    ybar,
legend style={
    at={(0.25,1)},
	anchor=north,legend columns=1},
    enlargelimits=0.3,
    ylabel={SHD},
    ylabel style={font=\Large},
    xlabel style={font=\Large},
	symbolic x coords={
	$\text{ideal}$,
	$\text{target}$,
	$\text{MVPC-PermC}$,
	$\text{MVPC-DRW}$,
	$\text{TD-PC}$},
    enlarge y limits=upper, ymin=0,
    xtick=data,
    bar width=10pt,
    width=1.8*\textwidth,
    axis x line*=bottom,
    axis y line*=left,
    yticklabel style = {font=\Large},
    x tick label style={font=\Large,rotate=30,anchor=east}
    ]
    \addplot +[black]
    [
    fill=green,
    postaction={
        pattern=crosshatch dots
    },
    pattern color=black,
    draw=black,
    error bars/.cd,
    y fixed,
    y dir=both, 
    y explicit,
    ]
    coordinates{
($\text{ideal}$,1.1)+-=(0,0.9944289)
($\text{target}$, 2.2)+-=(0,1.47573)
($\text{MVPC-PermC}$,5.1)+-=(0,3.510302)
($\text{MVPC-DRW}$,8)+-=(0,3.510302)
($\text{TD-PC}$,12.2)+-=(0,3.29309)
};
 \addplot
    [ 
    fill=blue,
    postaction={
    	pattern color=white,
        pattern=crosshatch dots
    },
    error bars/.cd,
    y fixed,
    y dir=both, 
    y explicit,
    ]
    coordinates{
($\text{ideal}$,1.1)+-=(0,0.9944289)
($\text{target}$, 1.7)+-=(0,1.828782)
($\text{MVPC-PermC}$,3.8)+-=(0,3.583915)
($\text{MVPC-DRW}$,7.5)+-=(0,3.583915)
($\text{TD-PC}$,12.2)+-=(0,3.29309)
};
 \legend{MNAR: no self-masking, MNAR: MAR + MCAR}

\end{axis}
  
\end{tikzpicture}

%% file: tikz/mnar_mech_f1.tex
\definecolor{orange}{RGB}{253,160,161}
\begin{tikzpicture}[scale=0.95]
  \begin{axis}[
    ybar,
legend style={
    at={(0.7,1)},
	anchor=north,legend columns=1},
    enlargelimits=0.3,
    ylabel={F$1$ score},
    ylabel style={font=\Large},
    xlabel style={font=\Large},
	symbolic x coords={
	$\text{ideal}$,$\text{target}$,$\text{MVPC-PermC}$,$\text{MVPC-DRW}$,$\text{test-wise\, PC}$},
    enlarge y limits=upper, ymin=0.8,
    xtick=data,
    bar width=10pt,
    width=1.8*\textwidth,
    axis x line*=bottom,
    axis y line*=left,
    yticklabel style = {font=\Large},
    x tick label style={font=\Large,rotate=30,anchor=east}
    ]
    \addplot +[black]
    [
    fill=green,
    postaction={
        pattern=crosshatch dots
    },
    pattern color=black,
    draw=black,
    error bars/.cd,
    y fixed,
    y dir=both, 
    y explicit,
    ]
    coordinates{
($\text{ideal}$,0.97470964)+-=(0,0.03135395)
($\text{target}$,0.96353246)+-=(0,0.02056618)
($\text{MVPC-PermC}$,0.94131148)+-=(0,0.04171016)
($\text{MVPC-DRW}$,0.89148505)+-=(0,0.03294932)
($\text{test-wise\, PC}$,0.851103)+-=(0,0.03710399)
};
 \addplot
    [ 
    fill=blue,
    postaction={
    	pattern color=white,
        pattern=crosshatch dots
    },
    error bars/.cd,
    y fixed,
    y dir=both, 
    y explicit,
    ]
    coordinates{
($\text{ideal}$,0.97470964)+-=(0,0.03135395)
($\text{target}$,0.97374442)+-=(0,0.02647145)
($\text{MVPC-PermC}$,0.96604278)+-=(0,0.03420483)
($\text{MVPC-DRW}$,0.91037014)+-=(0,0.04305899)
($\text{test-wise\, PC}$,0.85517074)+-=(0,0.02643475)
};
  \end{axis}
  
\end{tikzpicture}

%% file: tikz/mnar_NoV.tex
\definecolor{orange}{RGB}{253,160,161}
\begin{tikzpicture}[scale=1]
  \begin{axis}[
    ybar,
    legend style={
    draw=none,
    font=\Large,
    at={(0.89,1.2)},
	anchor=north,legend columns=5},
    enlargelimits=0.2,
    ylabel={SHD},
    ylabel style={font=\Large},
    xlabel={Number of Variables},
    xlabel style={font=\Large},
	symbolic x coords={20,50,100},
    enlarge y limits=upper, ymin=0.8,
    xtick=data,
    bar width=10pt,
    width=1.5 * \textwidth,
    axis x line*=bottom,
    axis y line*=left,
    yticklabel style = {font=\Large},
    x tick label style={font=\Large}
    ]
    \addplot +[black]
    [
    fill=yellow,
    postaction={
        pattern=crosshatch dots
    },
    pattern color=black,
    draw=black,
    error bars/.cd,
    y fixed,
    y dir=both, 
    y explicit,
    ]
    coordinates{
(20,3.1)+-=(0,3.28126)
(50,4.2)+-=(0,2.529822)
(100,16.3)+-=(0,5.45792)
};
 \addplot
    [ 
    fill=gray,
    postaction={
    	pattern color=white,
        pattern=crosshatch dots
    },
    error bars/.cd,
    y fixed,
    y dir=both, 
    y explicit,
    ]
    coordinates{
(20, 3.4)+-=(0,3.806427)
(50, 5.0)+-=(0,2.44949)
(100, 17)+-=(0,6.765928)
};
 \addplot +[black]
    [ 
    fill=orange,
    postaction={
        pattern=crosshatch dots
    },
    draw=black,
    error bars/.cd,
    y fixed,
    y dir=both, 
    y explicit,
    ]
    coordinates{

(20,5.8)+-=(0,3.224903)
(50,7.5)+-=(0,3.407508)
(100,13.1)+-=(0,4.483302)
};
\addplot +[black]
    [ 
    fill = cyan,
    postaction={
    	pattern color=black,
        pattern=south west lines
    },
    error bars/.cd,
    y fixed,
    y dir=both, 
    y explicit,
    ]
    coordinates{

(20, 8.7)+-=(0,3.224903)
(50,10.4)+-=(0, 3.407508)
(100,16.3)+-=(0,4.483302)

};
\addplot +[black]
    [ 
    pattern=south west lines,
    pattern color=black,
    draw=black,
    error bars/.cd,
    y fixed,
    y dir=both, 
    y explicit,
    ]
    coordinates{
(20,9.5)+-=(0,3.27448)
(50,15.0)+-=(0,4.921608)
(100,23.9)+-=(0,5.108816)
};
 \legend{ideal $\;$, target $\;$ , MVPC-PermC $\;$, MVPC-DRW $\;$, TD-PC $\;$}
\end{axis}
\end{tikzpicture}

%% file: tikz/mnar_NoV_f1.tex
\definecolor{orange}{RGB}{253,160,161}
\begin{tikzpicture}[scale=1]
  \begin{axis}[
    ybar,
    legend style={
    draw=none,
    font=\Large,
    at={(0.99,1.2)},
	anchor=north,legend columns=5},
    enlargelimits=0.2,
    ylabel={F1 score},
    ylabel style={font=\Large},
    xlabel={Number of Variables},
    xlabel style={font=\Large},
	symbolic x coords={20,50,100},
    enlarge y limits=upper, ymin=0.8,
    xtick=data,
    bar width=10pt,
    width=1.5 * \textwidth,
    axis x line*=bottom,
    axis y line*=left,
    yticklabel style = {font=\Large},
    x tick label style={font=\Large}
    ]
    \addplot +[black]
    [
    fill=yellow,
    postaction={
        pattern=crosshatch dots
    },
    pattern color=black,
    draw=black,
    error bars/.cd,
    y fixed,
    y dir=both, 
    y explicit,
    ]
    coordinates{
(20,0.95475400)+-=(0,0.01603516)
(50,0.97736993)+-=(0,0.02083407)
(100,0.96654568)+-=(0,0.03550662)
};
 \addplot
    [ 
    fill=gray,
    postaction={
    	pattern color=white,
        pattern=crosshatch dots
    },
    error bars/.cd,
    y fixed,
    y dir=both, 
    y explicit,
    ]
    coordinates{
(20, 0.95537567)+-=(0,0.01968621)
(50, 0.97036885)+-=(0,0.01558782)
(100, 0.97271344)+-=(0,0.03190197)
};
 \addplot +[black]
    [ 
    fill=orange,
    postaction={
        pattern=crosshatch dots
    },
    draw=black,
    error bars/.cd,
    y fixed,
    y dir=both, 
    y explicit,
    ]
    coordinates{
(20,0.97059160)+-=(0, 0.01027456)
(50,0.95682885)+-=(0, 0.02207709)
(100,0.92832555)+-=(0, 0.05097682)
};
\addplot +[black]
    [ 
    fill = cyan,
    postaction={
    	pattern color=black,
        pattern=south west lines
    },
    error bars/.cd,
    y fixed,
    y dir=both, 
    y explicit,
    ]
    coordinates{
(20,0.96104268)+-=(0,0.01442818)
(50,0.93674433)+-=(0,0.03210406)
(100,0.87246418)+-=(0,0.03838169)

};
\addplot +[black]
    [ 
    pattern=south west lines,
    pattern color=black,
    draw=black,
    error bars/.cd,
    y fixed,
    y dir=both, 
    y explicit,
    ]
    coordinates{
(20,0.94643288)+-=(0,0.01246147)
(50,0.92197365)+-=(0,0.02657883)
(100,0.8921976)+-=(0,0.0598558)

};
\end{axis}
\end{tikzpicture}

%% file: tikz/bin_mnar_mar.tex
\definecolor{orange}{RGB}{253,160,161}
\begin{tikzpicture}[scale=1]
  \begin{axis}
    [
    ybar,
    legend style={
    draw=none,
    font=\Large,
    at={(0.79,1.2)},
	anchor=north,legend columns=5},
    enlargelimits=0.4,
    ylabel={SHD},
    ylabel style={font=\Large},
    xlabel style={font=\Large},
	symbolic x coords={MAR, MNAR},
    enlarge y limits=upper, ymin=0.8,
    xtick=data,
    bar width=10pt,
    width=1.5 * \textwidth,
    axis x line*=bottom,
    axis y line*=left,
    yticklabel style = {font=\Large},
    x tick label style={font=\Large}
    ]
    \addplot +[black]
    [
    fill=yellow,
    postaction={
        pattern=crosshatch dots
    },
    pattern color=black,
    draw=black,
    error bars/.cd,
    y fixed,
    y dir=both, 
    y explicit,
    ]
    coordinates{
(MAR,4.3)+-=(0, 2.86937856222098)
(MNAR,4.3)+-=(0,2.86937856222098 )
};
 \addplot
    [ 
    fill=gray,
    postaction={
    	pattern color=white,
        pattern=crosshatch dots
    },
    error bars/.cd,
    y fixed,
    y dir=both, 
    y explicit,
    ]
    coordinates{
(MAR,5.7)+-=(0, 3.16403399335581)
(MNAR,5.6)+-=(0,2.98886823619465 )
};
 \addplot +[black]
    [ 
    fill=orange,
    postaction={
        pattern=crosshatch dots
    },
    draw=black,
    error bars/.cd,
    y fixed,
    y dir=both, 
    y explicit,
    ]
    coordinates{
(MAR,8.4)+-=(0, 4.42718872423573)
(MNAR,7.6)+-=(0,2.98886823619465 )
};
\addplot +[black]
    [ 
    fill = cyan,
    postaction={
    	pattern color=black,
        pattern=south west lines
    },
    error bars/.cd,
    y fixed,
    y dir=both, 
    y explicit,
    ]
    coordinates{
(MAR,6.1)+-=(0, 3.31494930412049)
(MNAR,8.0)+-=(0,3.2659863237109 )

};
\addplot +[black]
    [ 
    pattern=south west lines,
    pattern color=black,
    draw=black,
    error bars/.cd,
    y fixed,
    y dir=both, 
    y explicit,
    ]
    coordinates{
(MAR,13.2 )+-=(0,2.34757558155453)
(MNAR,14.4)+-=(0,3.59629438913631 )
};
 \legend{ideal $\;$, target $\;$ , MVPC-PermC $\;$, MVPC-DRW $\;$, TD-PC $\;$}
  \end{axis}
\end{tikzpicture}

%% file: tikz/bin_mnar_mar_f1.tex
\definecolor{orange}{RGB}{253,160,161}
\begin{tikzpicture}[scale=0.85]
  \begin{axis}[
    ybar,
    legend style={
    draw=none,
    font=\Large,
    at={(0.89,1.2)},
	anchor=north,legend columns=5},
    enlargelimits=0.4,
    ylabel={F1 score},
    ylabel style={font=\Large},
    xlabel style={font=\Large},
	symbolic x coords={MAR, MNAR},
    enlarge y limits=upper, ymin=0.7,
    xtick=data,
    bar width=10pt,
    width=1.5 * \textwidth,
    axis x line*=bottom,
    axis y line*=left,
    yticklabel style = {font=\Large},
    x tick label style={font=\Large}
    ]
    \addplot +[black]
    [
    fill=yellow,
    postaction={
        pattern=crosshatch dots
    },
    pattern color=black,
    draw=black,
    error bars/.cd,
    y fixed,
    y dir=both, 
    y explicit,
    ]
    coordinates{
(MAR,0.959664117558854 )+-=(0,0.0291904147692368)
(MNAR,0.959664117558854 )+-=(0,0.0291904147692368)
};
 \addplot
    [ 
    fill=gray,
    postaction={
    	pattern color=white,
        pattern=crosshatch dots
    },
    error bars/.cd,
    y fixed,
    y dir=both, 
    y explicit,
    ]
    coordinates{
(MAR,0.923044007768013 )+-=(0,0.0529694460005055)
(MNAR,0.927101190259085 )+-=(0,0.0310682994355919)
};
 \addplot +[black]
    [ 
    fill=orange,
    postaction={
        pattern=crosshatch dots
    },
    draw=black,
    error bars/.cd,
    y fixed,
    y dir=both, 
    y explicit,
    ]
    coordinates{
(MAR,0.860817835379239 )+-=(0,0.047493162484311)
(MNAR,0.863453679243153 )+-=(0,0.0539190494456379)
};
\addplot +[black]
    [ 
    fill = cyan,
    postaction={
    	pattern color=black,
        pattern=south west lines
    },
    error bars/.cd,
    y fixed,
    y dir=both, 
    y explicit,
    ]
    coordinates{
(MAR,0.879564276055504 )+-=(0,0.0500903586364823)
(MNAR,0.86018849455382 )+-=(0,0.0565192979681712)

};
\addplot +[black]
    [ 
    pattern=south west lines,
    pattern color=black,
    draw=black,
    error bars/.cd,
    y fixed,
    y dir=both, 
    y explicit,
    ]
    coordinates{
(MAR,0.809621279449921 )+-=(0,0.0336328536034787)
(MNAR,0.799614484242958 )+-=(0,0.0520714161527048)
};
\end{axis}
\end{tikzpicture}

%% file: tikz/CogUSA_barplot.tex
\begin{tikzpicture}[scale=0.5]
  \begin{axis}[
    ybar,
    bar width=0.6cm,
    enlargelimits=0.15,
    ylabel style={font=\Large},
    ylabel={Cost},
    symbolic x coords={MVPC, test-wise PC, list-wise PC, test-wise FCI, list-wise FCI},
    xtick=data,
    axis x line*=bottom,
    axis y line*=left,
    y tick label style={font=\Large},
    x tick label style={rotate=30,anchor=east, font=\Large},
    ]

        \addplot +[black]
    [
    fill=blue,opacity=0.5,
    postaction={
    	pattern color = white,
        pattern=crosshatch dots
    },
    pattern color=black,
    draw=black,
    error bars/.cd,
    error bar style={line width=2pt},
    y fixed,
    y dir=both, 
    y explicit,
    ]
    coordinates{
    (MVPC,2.564) +-= (0,0.870) 
    (test-wise PC,3.435) +-= (0,0.882)
    (list-wise PC,3.076)  +-= (0,0.718)
    (list-wise FCI,2.923) +-= (0,0.807)
    (test-wise FCI,2.846) +-= (0,0.874)};
  \end{axis}
\end{tikzpicture}

%% file: tikz/ATR_BMIetc.tex
\begin{tikzpicture}[
			scale=0.33,
            > = stealth, 
            shorten > = 1pt, 
            auto,
            node distance = 3cm, 
            semithick 
        ]

        \tikzstyle{every state}=[
            draw = black,
            thick,
            minimum size = 12mm
        ]
        \node[draw, ellipse]  (1) at  (0,4.5){ \small $Age$};
        \node[draw, ellipse] (2) at  (5,4.5) {\small $Gender$};
        \node[draw, ellipse] (3) at  (5,0) {\small $BMI$};
        \node[draw, ellipse] (4) at  (0,0) {\small $LSI$};
       \node[draw, ellipse] (6) at  (10,0) {\small $FAOS$};
        \path[-] (1) edge node {} (4);
        \path[-] (2) edge node {} (3);
    \end{tikzpicture}  

%% file: tikz/ATR_delete.tex
\begin{tikzpicture}[
			scale=0.3,
            > = stealth, 
            shorten > = 1pt, 
            auto,
            node distance = 3cm, 
            semithick 
        ]

        \tikzstyle{every state}=[
            draw = black,
            thick,
            minimum size = 12mm
        ]
        \node[draw, ellipse] (1) at  (0,5){\small $OP_{time}$};
        \node[draw, ellipse] (2) at  (6,5) {\small $FAOS$};
        \node[draw,circle]  (6) at  (3,0) {\small $S$};

		\path[-] (1) edge node {} (6);
        \path[-] (1) edge node {} (2);
        \path[-] (2) edge node {} (6);
    \end{tikzpicture} 

%% file: tikz/ATR_ours.tex
\begin{tikzpicture}[
			scale=0.3,
            > = stealth, 
            shorten > = 1pt, 
            auto,
            node distance = 3cm, 
            semithick 
        ]

        \tikzstyle{every state}=[
            draw = black,
            thick,
            minimum size = 12mm
        ]
        \node[draw, ellipse] (1) at  (0,5){\small $Op_{time}$};
        \node[draw, ellipse] (2) at  (6,5) {\small $FAOS$};
        \node[draw, circle] (6) at  (3,0) {\small $S$};

		\path[->] (1) edge node {} (6);
        \path[->] (2) edge node {} (6);
    \end{tikzpicture}

%% file: discussion.tex
\section{Discussion} \label{sec:discussion} 
In this work, we first provide a theoretical analysis about the influence of missing data on causal discovery under  Assumptions \ref{ass:re0}$\sim$\ref{ass:re2}. We then introduce two correction methods together with their conditions and a framework, Missing Value PC (MVPC). In this section, we provide an additional consideration of Assumption \ref{ass:re2}, no self-masking missingness, and the conditions of correction methods.

\paragraph{Violation of "no self-masking missingness."}
Although many works \citep{mohan2019aaaisymposium, mohan2018estimation} focus on the recoverability of SelF-masking Missingness (SFM), it is still a challenge for causal discovery in the presence of SFM. Moving forward to solve the SFM problem, we note that in linear Gaussian cases SFM does not affect MVPC, especially when the SFM indicator $R_x$ only has one direct cause $X$, such as in Figure \ref{fig:self-m-sing}. In this case, the result of the CI test of $X$ and $Y$ in test-wise deleted data implies the correct d-separation relation in the missingness graph. With the faithfulness assumption on the missingness graph,  we have $X\independent Y \Longleftrightarrow X\independent Y \mid R_x$; furthermore, under the faithful observability assumption, we have $X\independent Y\mid R_x \Longleftrightarrow X^* \independent Y \mid R_x=0$ which is what we test in the test-wise deleted data of $X$ and $Y$. Moreover, even though the test-wise deleted data may not follow the Gaussian distribution, 
the result of the linear Gaussian CI test is not influenced in practice.

\begin{figure}[!ht]%
\centering
\input{tikz/self_missing2.tex} 
\caption{ Self-masking missingness indicator with multiple direct causes: TD-PC produces an extra edge between $X$ and $Y$, but such self-masking missingness does not affect the other edges in the causal skeleton results, such as the edge between $X$ and $V_i\in \mathbf{V} \setminus \{X,Y\}$.}
\label{fig:self-m-mul}
\end{figure}
SFM affects MVPC results when the SFM indicator $R_x$ has multiple direct causes. For example, as the missingness graph in Figure \ref{fig:self-m-mul} shown, conditioning on the missingness indicator which is the direct common effect of two variables in a CI test produces an extraneous edge between them in the result given by MVPC. Removing such extraneous edges is challenging, because our correction methods are not applicable to the self-masking missingness scenario. 
More specifically, we cannot consistently estimate $P(R_x \mid X, Y)$ because $X$ is not available in observed data and 
\begin{align}\label{eq:self-eq}
    P(X, Y) = \frac{P(R_x =0 )}{ P(R_x = 0 \mid X, Y)} P(X, Y \mid R_x = 0). \nonumber
\end{align} 
However, such self-masking missingness indicator does not affect the other edges between $X$ and variables in $\mathbf{V} \setminus \{X,Y\}$ in the causal skeleton resulted by MVPC. Therefore, we specify in the output that edges between the self-masking variable and other direct causes of the self-masking missingness indicator are uncertain.

For further solving the SFM problem, a potential solution can be that one may propose a reasonable assumption of the missingness generation process regarding specific application information, and then identify the independence relation of $X$ and $Y$. For example, if considering $P(R_x=0 \mid X, Y)$ as a function of $X$ and $Y$, under mild assumption about the function form, we may identify the transformation between $P(X, Y \mid  R_x=0)$ and $P(X, Y)$ or even identify conditional independence relation of $X$ and $Y$ directly.

\paragraph{Comparison of two correction methods of MVPC.}
We propose a novel permutation-based correction method (PermC) for MVPC and provide the conditions of using this method. In Section \ref{sec:synthetic} we apply MVPC with PermC to the MNAR datasets satisfying the conditions and the datasets that do not satisfy its conditions. The similar results on these MNAR datasets with different generation methods indicate that the conditions of of PermC are not too strict such that most randomly generated causal graphs satisfy the conditions of PermC. Moreover, it is straightforward to extend the PermC to non-linear non-Gaussian cases. One could simply replace the linear regression method by a more complex model, and then follow the same procedure of PermC. As an example, we show how to extend such method for binary variable cases. Since the binary extension method is not in the linear Gaussian case, the condition becomes weaker as shown in Condition \ref{cond:corr1}$^*$.

We also introduce another correction method, density ratio weighted correction method (DRW). We implement it by applying kernel density estimation to the denominator and the numerator of the density ratio term in Equation \ref{eq:corr2}. This can be considered as the inverse probability weighted method \citep{horvitz1952generalization} for the missing data problem. Therefore, the condition of DRW is the condition of kernel density ratio estimation. In theory, DRW handles the MNAR cases that do not satisfy the conditions of PermC; however, the performance of DRW is limited by the kernel density estimation, which works well with low dimensional data (in our experiments the dimension should be lower than $4$). Moreover, DRW requires much more data than PermC to have the similar performance. Therefore, as shown in the experiments in Section \ref{sec:synthetic} MVPC with DRW performs relatively worse than MVPC with PermC under the same experiment setting.

%% file: tikz/self_missing2.tex
\begin{tikzpicture}[
		scale=0.4,
        > = stealth, 
        shorten > = 1pt, 
        auto,
        node distance = 3cm, 
        semithick 
    ]

    \tikzstyle{every state}=[
    	circle,
        draw = black,
        thick,
        minimum size = 8mm
    ]

    \node[state,fill = gray] (X) at  (0,0){$X$};
    \node[state, fill = white] (Y) at  (3,0) {$Y$};
    \node[state, fill = white] (V) at  (3,-3) {$V_i$};
	\node[state] (rx) at  (0,-3) {$R_x$};
    \path[->] (X) edge node {} (rx);
    \path[->] (Y) edge node {} (rx);
    \path[->] (Y) edge node {} (V);    
\end{tikzpicture}

%% file: conclusion.tex
\section{Conclusion}
\label{sec:discussion}
In this work, we address the problem of causal discovery in the presence of missing data. 
We first provide theoretical analysis to identify possible errors in the results given by a simple extension of PC. We then show that erroneous causal edges occur only in particular graph structures. 
Based on our analysis, we propose a framework Missing Value PC (MVPC) together with a novel correction method, permutation-based correction method, which corrects erroneous edges under mild assumptions. When the conditions of the permutation-based correction method are not satisfied, we provide the density ratio weighed correction method. For a better evaluation of causal discovery methods we develop a neuropathic pain diagnosis simulator, simulating real-world benchmark datasets. 
We then demonstrate the asymptotic correctness and the effectiveness of our method with the evaluation on the synthetic data, neuropathic pain diagnosis simulation data, and real-world applications.

As future work, we will explore the possibility of further relaxing the assumptions in MVPC, as well as work jointly with practitioners on causal analysis of large-scale healthcare applications in the presence of missing data.  